\documentclass{article} 
\usepackage{iclr2024_conference,times}
\usepackage{graphicx}

\usepackage{amsmath,amsfonts,bm}
\usepackage{amssymb}
\usepackage{bbm}

\usepackage{listings}
\usepackage{xcolor}
\usepackage{courier}

\definecolor{codegreen}{rgb}{0,0.6,0}
\definecolor{codegray}{rgb}{0.5,0.5,0.5}
\definecolor{codeblack}{rgb}{0.,0.,0.}
\definecolor{codepurple}{rgb}{0.58,0,0.82}
\definecolor{backcolour}{rgb}{0.95,0.95,0.92}

\lstdefinestyle{mystyle}{
    language=Python,
    backgroundcolor=\color{backcolour},   
    commentstyle=\color{codegreen},
    keywordstyle=\color{codeblack},
    numberstyle=\tiny\color{codegray},
    stringstyle=\color{codepurple},
    basicstyle=\ttfamily\footnotesize,
    breakatwhitespace=false,         
    breaklines=true,                 
    captionpos=b,                    
    keepspaces=true,                   
    showspaces=false,                
    showstringspaces=false,
    showtabs=false,                  
    tabsize=2,
    aboveskip=0pt,
    belowskip=-3pt,
    columns=fullflexible,
    literate={-}{-}1
}

\makeatletter
\def\lst@outputspace{{\ifx\lst@bkgcolor\empty\color{white}\else\lst@bkgcolor\fi\lst@visiblespace}}
\makeatother
\usepackage{amsthm}
\newtheorem{theorem}{Theorem}
\newtheorem{corollary}{Corollary}[theorem]
\newtheorem{proposition}{Proposition}

\def\1{\bm{1}}

\def\cst{{\rm cst}}
\newcommand{\rmat}[2]{\mathcal{M}_{#1,#2}(\mathbb{R})}

\DeclareMathOperator{\spn}{span}

\DeclareMathOperator{\Tr}{Tr}
\newcommand{\Trp}[1]{\Tr\left(#1\right)}
\DeclareMathOperator{\eigvec}{eigvec}

\def\vv{{\bm{v}}}

\def\vx{{\bm{x}}}

\def\mA{{\bm{A}}}
\def\mB{{\bm{B}}}

\def\mD{{\bm{D}}}

\def\mG{{\bm{G}}}
\def\mH{{\bm{H}}}
\def\mI{{\bm{I}}}

\def\mM{{\bm{M}}}

\def\mP{{\bm{P}}}

\def\mS{{\bm{S}}}

\def\mU{{\bm{U}}}
\def\mV{{\bm{V}}}
\def\mW{{\bm{W}}}
\def\mX{{\bm{X}}}
\def\mY{{\bm{Y}}}
\def\mZ{{\bm{Z}}}

\def\mSigma{{\bm{\Sigma}}}

\DeclareMathAlphabet{\mathsfit}{\encodingdefault}{\sfdefault}{m}{sl}
\SetMathAlphabet{\mathsfit}{bold}{\encodingdefault}{\sfdefault}{bx}{n}

\DeclareMathOperator*{\argmax}{arg\,max}
\DeclareMathOperator*{\argmin}{arg\,min}

\usepackage{hyperref}
\usepackage{url}
\usepackage{amssymb}
\usepackage{bbm}
\usepackage{cleveref}

\title{
Learning by Reconstruction Focuses on Uninformative Features for Perception
}

\author{Antiquus S.~Hippocampus, Natalia Cerebro \& Amelie P. Amygdale \thanks{ Use footnote for providing further information
about author (webpage, alternative address)---\emph{not} for acknowledging
funding agencies.  Funding acknowledgements go at the end of the paper.} \\
Department of Computer Science\\
Cranberry-Lemon University\\
Pittsburgh, PA 15213, USA \\
\texttt{\{hippo,brain,jen\}@cs.cranberry-lemon.edu} \\
\And
Ji Q. Ren \& Yevgeny LeNet \\
Department of Computational Neuroscience \\
University of the Witwatersrand \\
Joburg, South Africa \\
\texttt{\{robot,net\}@wits.ac.za} \\
\AND
Coauthor \\
Affiliation \\
Address \\
\texttt{email}
}

\begin{document}

\maketitle

\begin{abstract}
Input space reconstruction is an attractive representation learning paradigm. Despite interpretability of the reconstruction and generation, we identify a misalignment between learning by reconstruction, and learning for perception. We show that the former allocates a model's capacity towards a subspace of the data explaining the observed variance--a subspace with uninformative features for the latter. 
For example, the supervised TinyImagenet task with images projected onto the top subspace explaining 90\% of the pixel variance can be solved with 45\% test accuracy. Using the bottom subspace instead, accounting for only 20\% of the pixel variance, reaches 55\% test accuracy.
%
%
The features for perception being learned last explains the need for long training time, e.g., with Masked Autoencoders. 
Learning by denoising is a popular strategy to alleviate that misalignment. We prove that while some noise strategies such as masking are indeed beneficial, others such as additive Gaussian noise are not. Yet, even in the case of masking, we find that the benefits vary as a function of the mask's shape, ratio, and the considered dataset. While tuning the noise strategy without knowledge of the perception task seems challenging, we provide first clues on how to detect if a noise strategy is never beneficial regardless of the perception task.
\end{abstract}

\twocolumn[
\icmltitle{Learning by Reconstruction Produces Uninformative Features For Perception}



\icmlsetsymbol{equal}{*}

\begin{icmlauthorlist}
\icmlauthor{Randall Balestriero}{yyy}
\icmlauthor{Yann LeCun}{comp}
\end{icmlauthorlist}

\icmlaffiliation{yyy}{Independent}
\icmlaffiliation{comp}{NYU}

\icmlcorrespondingauthor{Randall Balestriero}{randallbalestriero@gmail.com}

\icmlkeywords{Machine Learning, ICML}

\vskip 0.3in
]



\printAffiliationsAndNotice{} 

\section{Introduction}
\label{sec:intro}

\begin{figure}[t!]
    \centering\includegraphics[width=\linewidth]{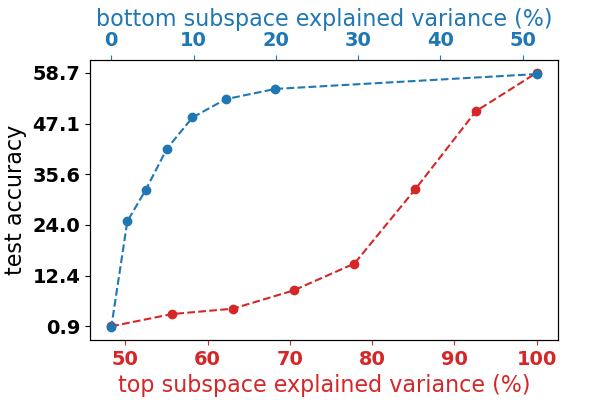}\\
    \begin{minipage}{0.49\linewidth}
        \centering
        \color{blue}bottom 25\% variance
    \includegraphics[width=\linewidth]{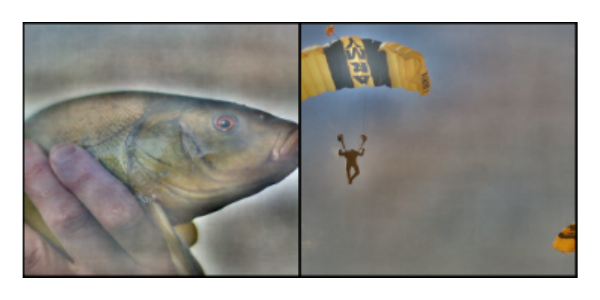}
    \end{minipage}
    \begin{minipage}{0.49\linewidth}
        \centering
        \color{red}top 75\% variance
    \includegraphics[width=\linewidth]{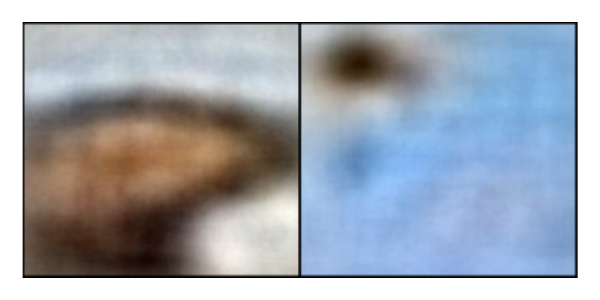}
    \end{minipage}\\
    \vspace{-0.2cm}
    \includegraphics[width=\linewidth]{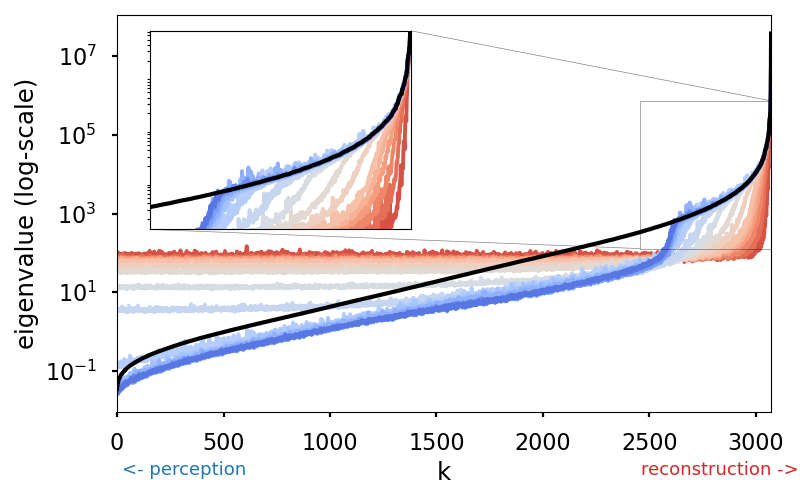}
    \vspace{-0.5cm}
    \caption{\small {\bf Features for reconstruction are uninformative for perception (top):} TinyImagenet ResNet9 top-1 accuracy when trained and validated on images projected on the top-subspace (red) or bottom subspace (blue) of explained variance, corresponding images displayed in the {\bf middle} and in \cref{fig:images_pca}. {\bf Perception features are learned last (bottom):} training loss evolution (red to blue) of reconstructed training images from a deep Autoencoder projected onto the eigenspace of the original data (black). The top eigenspace (right) is learned first, and then, if training lasts long enough, the features most useful for perception (left) are finally learned. This explains why learning by performances on perception task keep increasing long after reconstructed samples look appealing.}
    \label{fig:teaser}
\end{figure}

One of the far reaching mandate of deep learning is to provide a self-contained methodology to learn intelligible and universal representations of data \cite{lecun2015deep}. That is, to learn a nonlinear transformation of the data producing a parsimonious and informative representation which can be used to solve numerous downstream tasks. Significant progress has been made through the lens of supervised learning, i.e., by learning a representation that maps the observed data to provided labels of an priori known downstream task \cite{krizhevsky2012imagenet}. Labels being costly and over-specialized, much progress was also made through the lens of unsupervised learning \cite{barlow1989unsupervised,ghahramani2003unsupervised} roughly falling into three camps. First, reconstruction-based methods that produce (compressed) latent representations of the data nonetheless sufficient to recover most of the original data, e.g., Denoising/Variational/Masked Auto-Encoders \cite{vincent2010stacked,kingma2013auto,he2022masked} and Autoregressive models \cite{van2016conditional,chen2018pixelsnail}. Second, score matching which is often solved by setting up a surrogate supervised task of classifying observed samples from noise \cite{hyvarinen2005estimation}. Third, Self-Supervised Learning (SSL) \cite{chen2020big,zbontar2021barlow,bardes2021vicreg,balestriero2023cookbook}--whose contrastive methods can be thought of as generalized versions of score matching and Noise Contrastive Estimation \cite{gutmann2010noise}--combining an invariance term bringing together representations of data known to be semantically similar, or generated to be so, and an anti-collapse term making sure that not all representations become similar.

In recent years, SSL emerged as the preferred solution--regularly reaching new state-of-the-arts through careful experimental design \cite{garrido2023rankme}. Yet, reconstruction-based methods maintain a large presence due to their ability to provide reconstructed samples that are human interpretable, enabling informed quality assessment of a model \cite{selvaraju2016grad,bordes2021high,brynjolfsson2023generative,baidoo2023education}. Despite that benefit, reconstruction-based learning falls behind SSL as it requires fine-tuning to reach the state-of-the-art. One of the most popular reconstruction-based learning strategy that emerged as the solution of choice in recent years is the Masked-Autoencoder \cite{he2022masked}.

We ask the following question: {\em Why reconstruction based learning easily produces compelling reconstructed samples but fails to provide competitive latent representations for perception?} We can pinpoint that observation to at least three reasons.

\textbf{R1: Misaligned.}~The features with most reconstructive power are the least informative for perceptual tasks as depicted in \cref{fig:teaser} (top). Instead, images projected onto the bottom subspace remain informative for perception.

\textbf{R2: Ill-conditioned.}~Features useful for perception (low variance subspace) are learned last as depicted in \cref{fig:teaser} (bottom), in favor of learning first the top subspace of the data which explains most of the pixel variance but fails to solve perceptual tasks.

\textbf{R3: Ill-posed.}~There exists different model parameters producing the same train and test reconstruction error but exhibiting significant performance gaps for perceptual tasks as depicted in \cref{fig:compare}, where for a given reconstruction error the top-1 accuracy on Imagenet-10 can vary from 50\% to almost 90\%.

The findings from R1, R2, and R3 provide first clues as to why learning by reconstruction requires long training time and fine-tuning. Yet, those findings alone do not answer the following question: {\em Why Masked Autoencoders were able to provide a significant improvement in the quality of the learned representation for solving perception tasks?} We will prove that the hindering R1, R2 and R3, can be pushed back through careful design of the noise distribution used in denoising autoencoders. In particular, we will demonstrate that masking is provably beneficial while other noise distributions, e.g., additive Gaussian noise, are not. We hope that our findings will help skew further research in learning by reconstruction to explore alternative noise distributions as they are the main driver of learning useful representations for perception.

\section{Background and Notations}
\label{sec:background}

{\bf Notations:}~We denote by $\vx_n \in \mathbb{R}^{D}, n=1,\dots,N$ the $n^{\rm th}$ input sample, e.g., a $(H, W, C)$ image flattened to a $D=H \times W \times C$ dimensional vector. The entire training set is collected into the matrix $\mX\triangleq [\vx_1,\dots,\vx_N] \in \rmat{D}{N}$ where $\rmat{n}{m}$ is the vector space of real $n \times m$ matrices. Throughout our study, vectors will always be column-vectors, and matrices are built by horizontally stacking column-vectors, i.e., they are column major. Unless specified otherwise, we assume that the input matrix $\mX$ is full-rank. In practice, if this is not the case, one can easily disregard the subspace associated with $0$ singular values and apply our analysis on that filtered matrix instead.

\noindent
{\bf Learning by reconstruction.}~Learning representations by fitting a model's parameters $\theta$ to produce a reconstruction of presented inputs as in
\begin{align}
    \min_{\theta} \mathbb{E}_{\vx \sim p_{\vx}}\left[d\left(g_{\theta}\left(f_{\theta}(\vx)\right), \vx\right)\right],\label{eq:AE}
\end{align}
is common \cite{bottou2012stochastic,kingma2014adam,lecun2015deep}. The reconstruction provides a qualitative signal enabling one to easily assess the quality of the model, and even interpret trained classification models \cite{zeiler2014visualizing,mahendran2015understanding,olah2017feature,shen2020interpreting}. In its simplest form, the encoder $f_{\theta}:\mathbb{R}^{D}\mapsto \mathbb{R}^{K}$ and the decoder $g_{\theta}:\mathbb{R}^{K}\mapsto \mathbb{R}^{D}$ are linear, possibly with shared parameters. In such settings, the optimal parameters are obtained from Principal Component Analysis \cite{wold1987principal}. Many variants of \cref{eq:AE} have emerged, such as denoising and masked autoencoders (MAEs)\cite{vincent2010stacked,he2022masked}. The objective remains similar: learn a low-dimensional latent embedding of the data that is able to reconstruct the original samples while being robust to some noise perturbation added onto the samples as in
\begin{align}
    \min_{\theta} \mathbb{E}_{\vx \sim p_{\vx}}\left[\mathbb{E}_{\vx' \sim p_{\vx'|\vx}}\left[d\left(g_{\theta}\left(f_{\theta}(\vx')\right), \vx\right)\right]\right],\label{eq:DAE}
\end{align}
with $p_{\vx'|\vx}$ applying some (conditional) noise transformation to the original input, e.g., $\vx' \sim \mathcal{N}(\vx,\epsilon \mI), \epsilon>0$.

\noindent
{\bf Known limitations.}~
Learning by reconstruction is widely popular and thus heavily studied. Major axes of research evolve around (i) deriving novel loss functions for specific datasets that better align with semantic distance \cite{wang2004image,kulis2013metric,balle2016end,bao2017cvae}, (ii) explaining the learned embedding dimensions \cite{tran2017disentangled,esmaeili2019structured,mathieu2019disentangling}, and (iii) imposing structure in the embedding space such as clustered embeddings \cite{jiang2016variational,dilokthanakul2016deep,lim2020deep,seydoux2020clustering}. Despite the rich literature, the current solutions of choice to learn representation in computer vision still rely on the Mean Squared Error loss in pixel space with the possible application of structured noise ($p_{\vx'|\vx}$ in \cref{eq:DAE}), e.g., as employed by the current state-of-the-art solution of masked-autoencoders (MAEs) \cite{he2022masked}. Yet, even that solution learns a representation that needs to be fine-tuned to compete with state-of-the-art, e.g., MAE's performances are drastically determined by two parameters (i) the training time, i.e., evaluation performance of the learned representation does not plateau even after 1600 epochs on Imagenet, and (ii) the need to fine tune, i.e., evaluation performance with and without fine tuning have a significant gap going from 70\% to 84\% on top-1 Imagenet classification task.

Our study will propose some first hints as to why even current state-of-the-art solutions are poised with slow training, and the need to fine-tune (\cref{sec:theory}). We will conclude by proving that MAEs's masking strategy partially alleviates those limitations (\cref{sec:guidance}), showing that the most rewarding findings for learning by reconstruction may emerge from novel denoising strategies.

\section{Rich Features for Reconstruction are Poor Features For Perception}
\label{sec:theory}

This section provides the theoretical ground and empirical validation of {\bf R1} and {\bf R2} from \cref{sec:intro}, namely, that learning by reconstruction learns features that are misaligned with common perception tasks. We start by deriving a closed form alignment measure between those two tasks in \cref{sec:linear} and conclude by empirically measuring that mismatch in \cref{sec:misfit}.

\subsection{How To Measure The Alignment Between Reconstruction and Supervised Tasks}
\label{sec:linear}

As a starting point to our study, we will build intuition and obtain theoretical results in the linear regime. As we will see at the end of this \cref{sec:linear}, this seemingly simplified setting turns out to be informative of practical cases.

Let's consider an encoder mapping $\mV \in \rmat{K}{D}$, a decoder mapping $\mZ \in \rmat{D}{K}$, and a predictor head $\mW \in \rmat{C}{K}$, where $C$ is the number of target dimensions, or classes. The targets for $\mX\in \rmat{D}{N}$ are given by $\mY \in \rmat{D}{N}$. The combination of the supervised and reconstruction losses is given by
\begin{align}
    \mathcal{L}(\mV\hspace{-0.07cm},\mW\hspace{-0.07cm},\mZ) \hspace{-0.07cm}= &\| \mW^\top\mV^\top\hspace{-0.04cm}\mX \hspace{-0.07cm}- \hspace{-0.07cm}\mY\|_F^2
    + \lambda\hspace{-0.02cm}\|\mZ^\top\mV^\top\hspace{-0.07cm}\mX \hspace{-0.07cm}-\hspace{-0.07cm} \mX \|_F^2,\label{eq:bilinear}
\end{align}
where the latent representation, $\mV\mX$, is shared between the two losses, $\lambda \geq 0$ controls the trade-off between the two terms. Quantifying how the optimal parameters of \cref{eq:bilinear} vary with $\lambda$ will be key to assess how the two losses are aligned. As a starting point, let's formalize below the optimal parameters of this loss function using the notations $\mM = \mP_{\mM}\mD_{\mM}\mP_{\mM}^\top$ for the eigendecomposition of a symmetric semi-definite positive matrix $\mM$. We will also denote, to lighten notations, $\mA \triangleq \mX\left(\mY^\top\mY + \lambda \mX^\top\mX\right)\mX^\top$.

\begin{figure*}[t!]
\centering
    \includegraphics[width=0.8\linewidth]{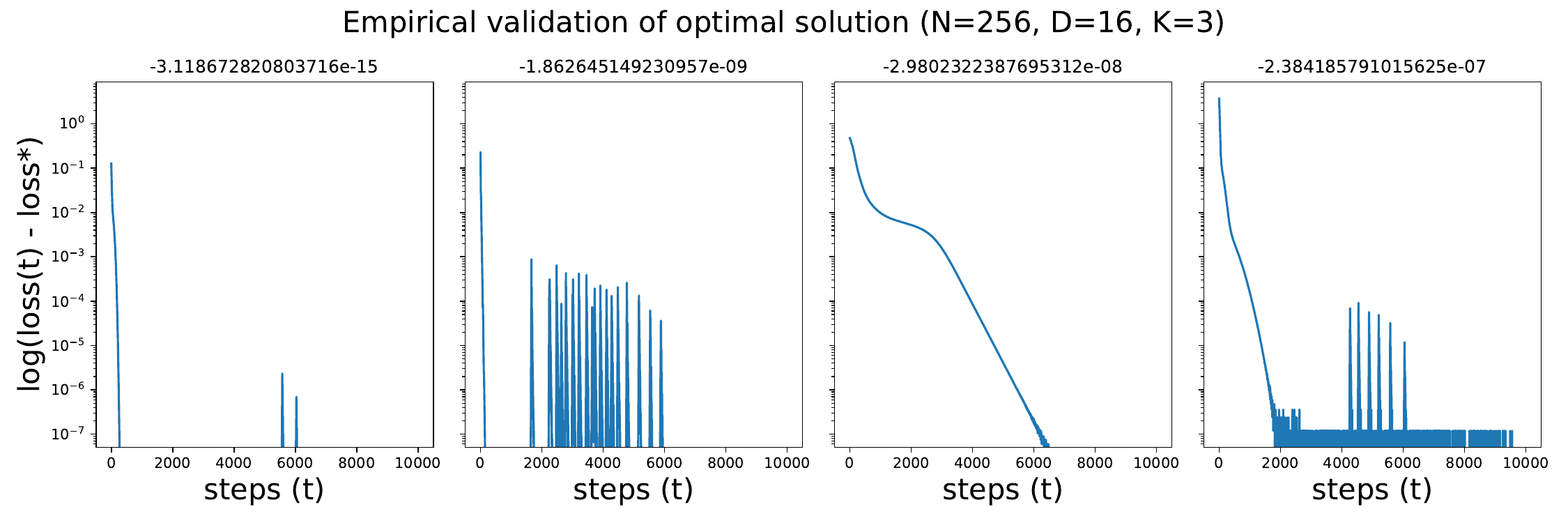}
    \includegraphics[width=0.8\linewidth]{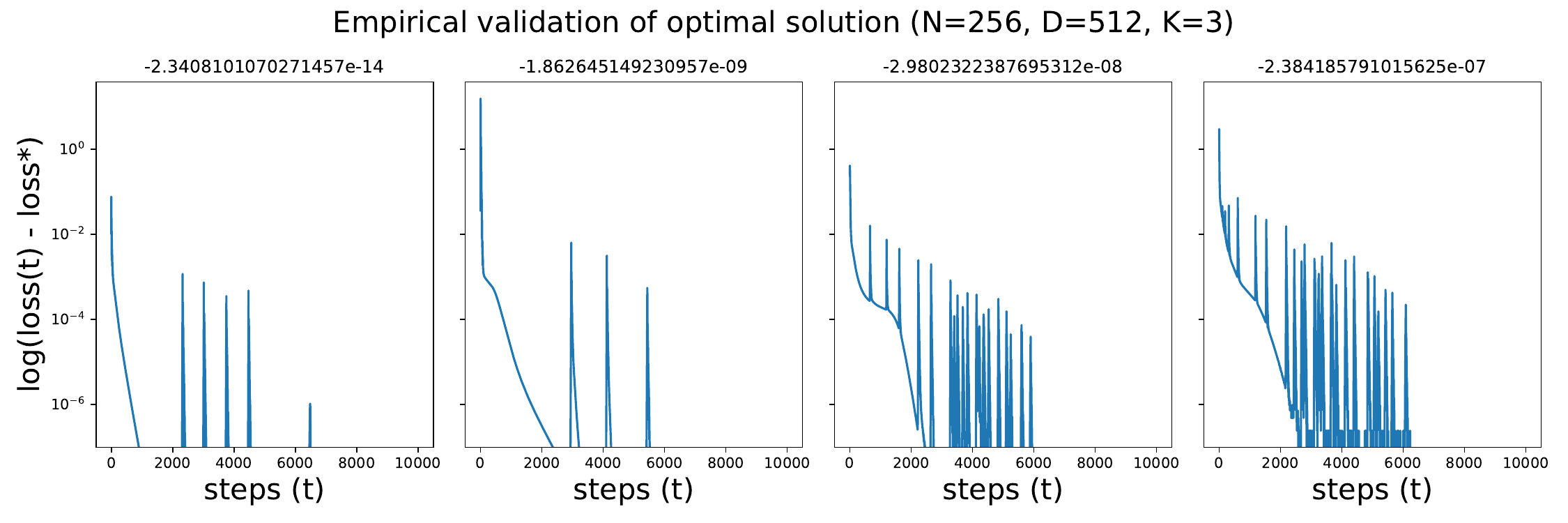}
    \includegraphics[width=0.8\linewidth]{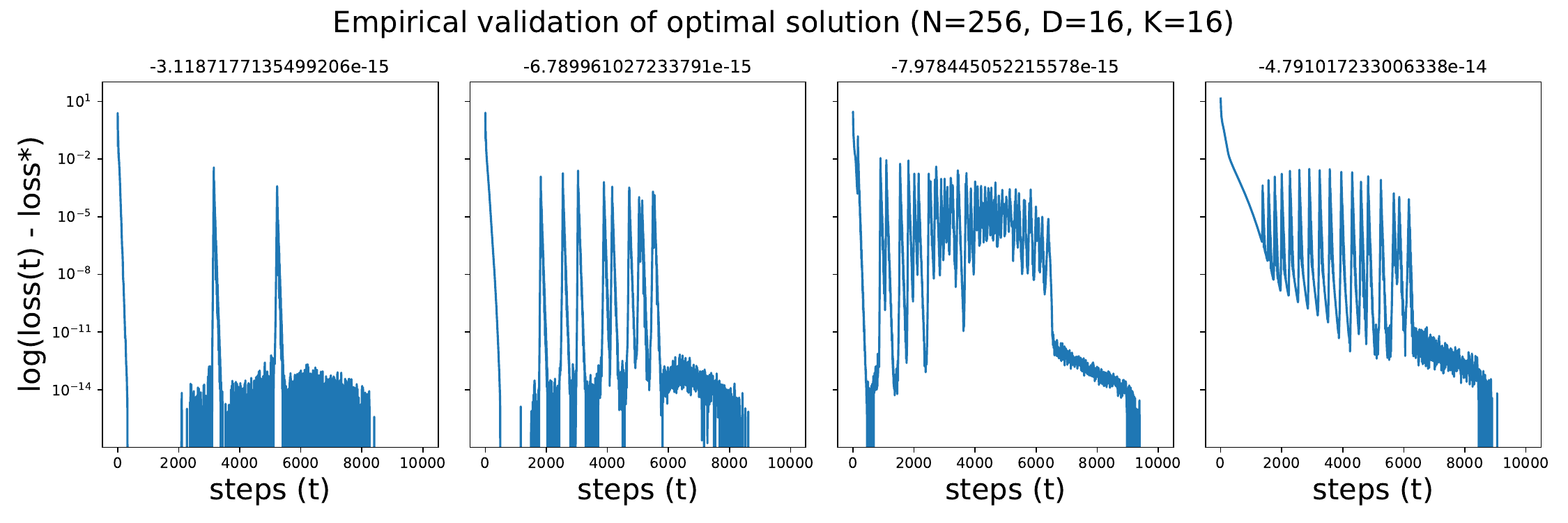}
    \includegraphics[width=0.8\linewidth]{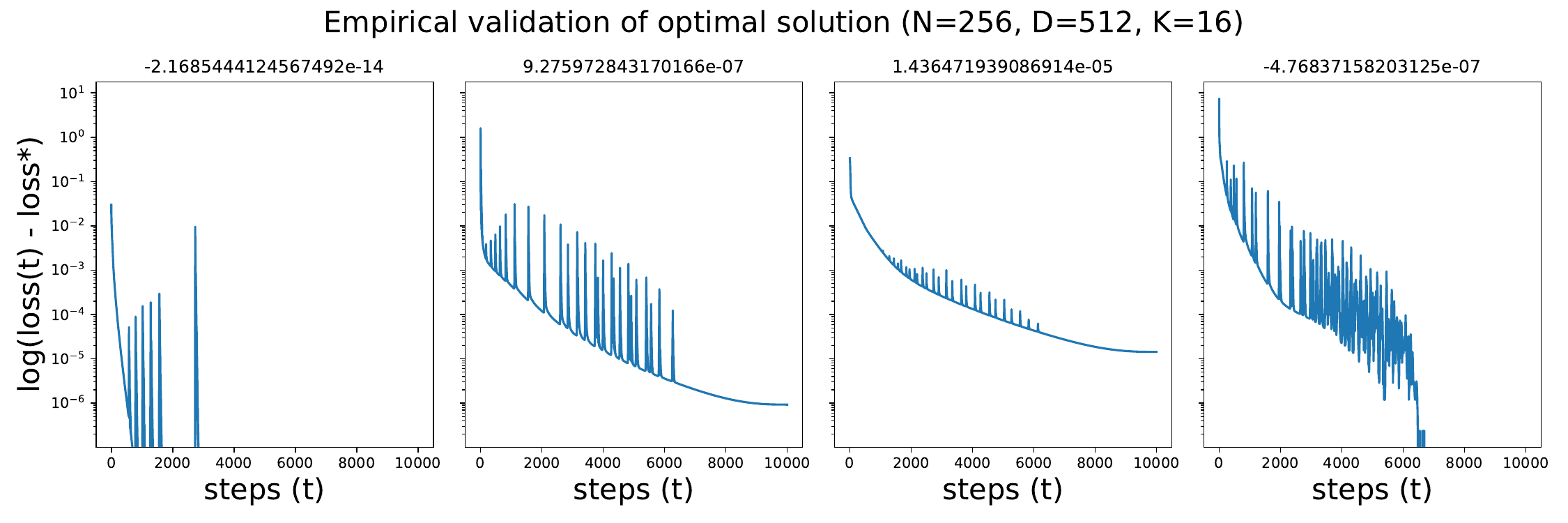}
    \caption{Empirical validation of \cref{thm:linear_solution} comparing the loss value at the optimum (from \cref{eq:V,eq:W,eq:Z}) against the one minimized with gradient descent (Adam optimizer) ({\bf y-axis}) during gradient steps (t, {\bf x-axis}). We expect that different to get close to $0$ as the gradient updates converge to the minimum value of the loss. Although that quantity (loss(optimum) - loss(t)) is nonnegative in theory, we observe that its minimum value (reported in the title of each subplot) is sometimes negative with negligible value due to round off error. We compare numerous values of $K,D,N$ as given in the titles of each {\bf row}, and different values of $\lambda \in \{0.0, 0.1, 1, 10\}$ ({\bf column}).}
    \label{fig:validation_general}
\end{figure*}

\begin{theorem}
\label{thm:linear_solution}
    The loss function from \cref{eq:bilinear} is minimized for
 \begin{align}
    \mV^*  &\text{ spans } \mP_{\mX\mX^\top}\mD^{-\frac{1}{2}}_{\mX\mX^\top}(\mP_{\mH})_{.,1:K},\label{eq:V}\\
    \mW^* =& \left({\mV^*}^\top\mX\mX^\top\mV^*\right)^{-1}{\mV^*}^\top\mX\mY^\top,\label{eq:W}\\
    \mZ^* =& \left({\mV^*}^\top\mX\mX^\top\mV^*\right)^{-1}{\mV^*}^\top\mX\mX^\top,\label{eq:Z}
\end{align}
where $\mH \triangleq  \mD^{-\frac{1}{2}}_{\mX\mX^\top}\mP^\top_{\mX\mX^\top}\mA\mP_{\mX\mX^\top} \mD^{-\frac{1}{2}}_{\mX\mX^\top}$. (Proof in \cref{proof:linear_solution}, empirical validation in \cref{fig:validation_general}.)
\end{theorem}

We observe that the optimal solutions from \cref{thm:linear_solution} continuously interpolates between the standard least square (OLS) problem ($\lambda=0$) and the unsupervised linear autoencoder or Principal Component Analysis (PCA) setting ($\lambda \to \infty$). We formalize below that we recover the optimal solutions for each of those extreme cases from \cref{thm:linear_solution}.

\begin{corollary}
\label{thm:PCA}
The solution from \cref{thm:linear_decoder} recovers the OLS solution for ${\mW^*}^\top{\mV^*}^\top$ as $\lambda \to 0$, and the PCA solution for ${\mZ^*}^\top{\mV^*}^\top$ as $\lambda \to \infty$. (Proof in \cref{proof:PCA}.)
\end{corollary}

That observation should comfort the reader that \cref{eq:bilinear} accurately conveys both ends of the spectrum from supervised learning to reconstruction based learning, while continuously interpolating in-between. 

{\bf Condition for perfect alignment.}~The result from \cref{thm:linear_solution} enables us to formalize the condition for perfect alignment between the two tasks, i.e., under which condition the solution $\mV^*$ is not impacted by $\lambda$. 

\begin{proposition}
\label{thm:condition}
    The supervised and reconstruction tasks are aligned (the optimal solutions do not depend on $\lambda$) iff the intersection of the top-$K$ eigenspaces of $\mX^\top\mX$ and $\mY^\top\mY$ is of dimension $K$.
\end{proposition}

In other words, whenever the condition in \cref{thm:condition} holds, the matrix $\mP_{\mH}$ (recall \cref{thm:linear_solution}) will include the same eigenvectors (up to rotation) for any $\lambda$, making the optimal parameters (\cref{eq:V,eq:W,eq:Z}) independent of $\lambda$. In practice, we will see that \cref{thm:condition}'s alignment condition is never fulfilled, pushing the need to define a more precise measure of alignment between the two tasks.

\begin{figure}[t!]
    \centering
    \includegraphics[width=\linewidth]{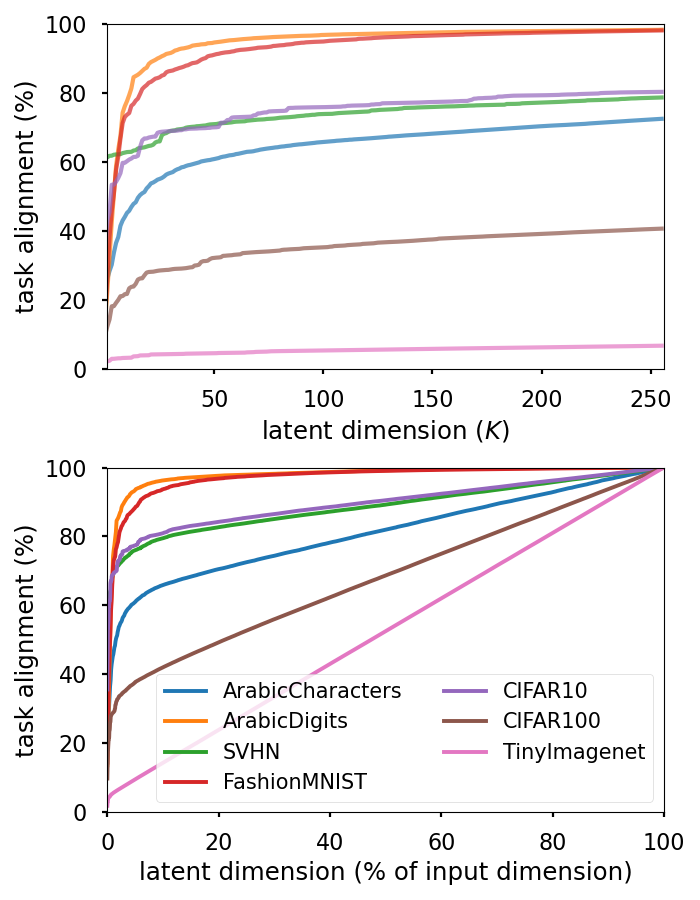}
    \caption{Depiction of the closed form alignment measure from \cref{eq:alignment} measuring the minimum supervised training error achievable given the optimal reconstruction parameters, as per \cref{thm:linear_solution,thm:alignment}. {\bf Top:} depiction in term of the latent dimension $K$ (x-axis). {\bf Bottom:} depiction in term of the ratio of the latent dimension $K$ to the input dimension $D$. We clearly observe that as the dataset becomes more realistic (going from background-free images to CIFAR and then to TinyImagenet), as the alignment between the reconstruction and supervised task lessens. In particular, when going to TinyImagenet, we observe that the alignment only increases linearly with respect to the latent space dimension.}
    \label{fig:alignment}
\end{figure}

\begin{figure*}[t!]
    \centering
    \begin{minipage}{0.49\linewidth}
        \centering
        CIFAR-DC-depth3\\
        \includegraphics[width=\linewidth]{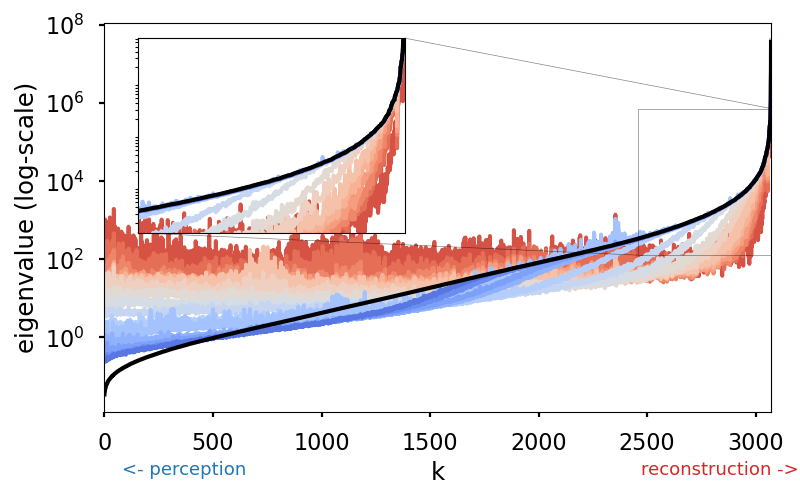}
    \end{minipage}
    \begin{minipage}{0.49\linewidth}
        \centering
        CIFAR-DC-depth5\\
        \includegraphics[width=\linewidth]{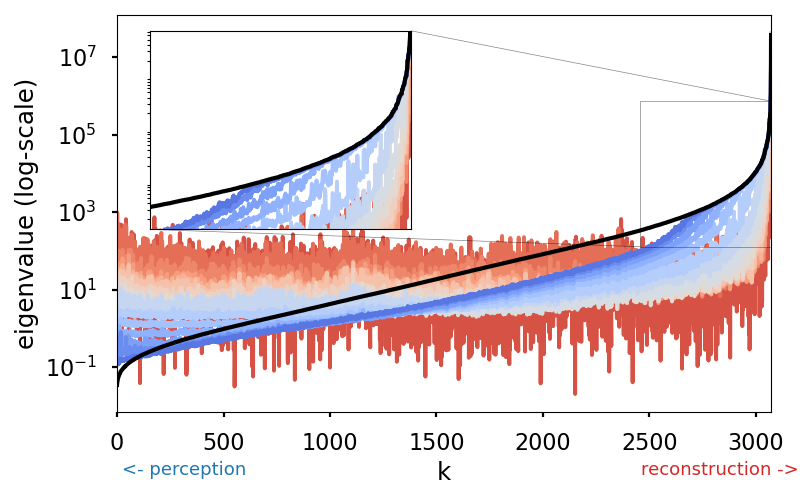}
    \end{minipage}\\
    \begin{minipage}{0.49\linewidth}
        \centering
        CIFAR-MLP-depth3\\
        \includegraphics[width=\linewidth]{evolution/spectrum_evolution_cifar10_MLPAE_3.png}
    \end{minipage}
    \begin{minipage}{0.49\linewidth}
        \centering
        CIFAR-MLP-depth5\\
        \includegraphics[width=\linewidth]{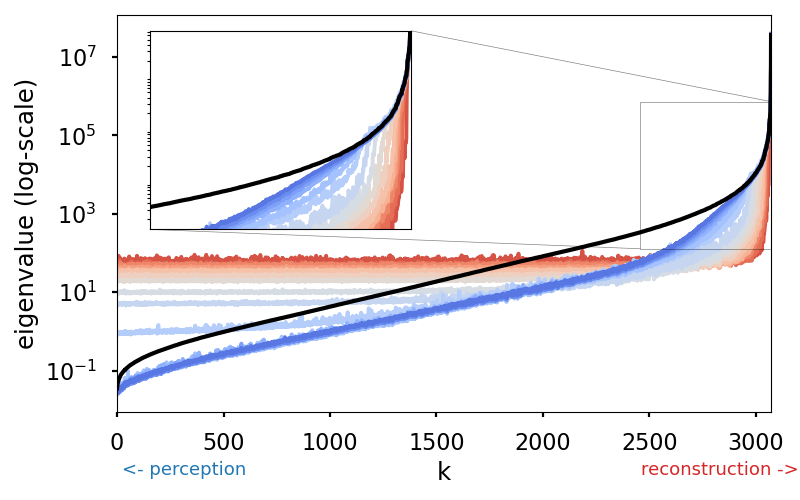}
    \end{minipage}
    \caption{Reprise of \cref{fig:teaser} for additional autoencoder architectures: convolutional encoder and deconvolutional decoder ({\bf top}) and MLP encoder and decoder ({\bf bottom}). We clearly observe that the top subspace is learned first during training, which is the one that best minimize the reconstruction loss but that contains the least informative features for perception, as per \cref{fig:classification}.}
    \label{fig:extra_focus}
\end{figure*}

{\bf Continuous measure of alignment.}~
As we aim to measure the tasks alignment more precisely that in a yes/no setting (\cref{thm:condition}), we propose the following continuous measure
\begin{align}
    \text{alignment}(k) \triangleq \frac{\| \mY^\top\mY(\mP_{\mX\mX^\top})_{1:k}\|_F^2}{\|\mY^\top\mY\mP_{\mX\mX^\top}\|_F^2},\label{eq:alignment}
\end{align}
where $\|\mY^\top\mY\mP_{\mX\mX^\top}\|_F^2$ simplifies as $\|\mY^\top\mY\|_F^2$ when $D=N$. We assume that the supervised task can be at least partially solved from $\mX$, ensuring $\|\mY^\top\mY\mP_{\mX\mX^\top}\|_F^2 > \epsilon$. In words, \cref{eq:alignment} is the (scaled) minimum supervised training error that can be achieved given the representation $(\mV^\top\mX)$ minimizing the reconstruction loss, which is measured by how much of the matrix $\mY^\top\mY$ can be reconstructed from the top-$k$ subspace of $\mX^\top\mX$, as formalized below.

\begin{corollary}
    \label{thm:alignment}
    $\text{alignment}(k)$ from \cref{eq:alignment} increases with $k$, has value $0$ iff the two losses are misaligned, and has value $1$ iff the two losses are aligned. (Proof in \cref{proof:alignment}.)
\end{corollary}

Although the measure from \cref{eq:alignment} is motivated from the linear setting of \cref{thm:linear_solution}, we will demonstrate at the end of this \cref{sec:linear} that it actually aligns well with practical settings.

{\bf Findings.}~We now propose to evaluate the closed form alignment metric (\cref{eq:alignment}) on a few datasets. Note that it can be implemented efficiently as detailed in \cref{sec:fast_alignment}. In \cref{fig:alignment}, we measure the metric of \cref{eq:alignment} for a sweep of latent dimension $K$ over $7$ different datasets. We observe three striking regimes. First, for images without background, reconstruction and classification tasks are very much aligned even for small latent dimension, as low as $20\%$ of the input dimension. Second, when comparing datasets with same image distributions but different number of classes (CIFAR10 to CIFAR100) the misalignment increases between the two task, especially for small embedding dimension. This follows our intuition that additional budget must be devoted to separate more classes. And as the subspace of the data used for reconstruction and classification do not align (recall \cref{thm:condition}), a greater misalignment is measured. And third, when looking at more realistic images (higher resolution and more diverse) such as tiny-imagenet, we observe that the alignment only increases linearly with the latent space dimension $K$, requiring $K=D$ in that case to ensure alignment. We thus conclude that {\em the presence of background, finer classification tasks, and higher resolution images, are all factors that drastically decrease the alignment between learning features for perception tasks and learning features that reconstruct those images.}

\begin{figure*}[t!]
    \centering
    \includegraphics[width=\linewidth]{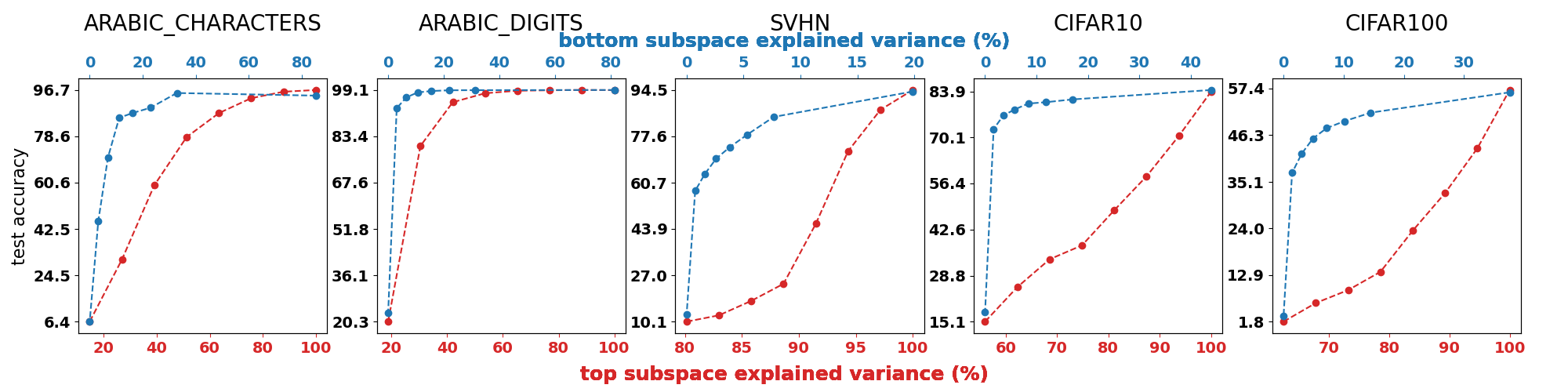}\\
    \includegraphics[width=\linewidth]{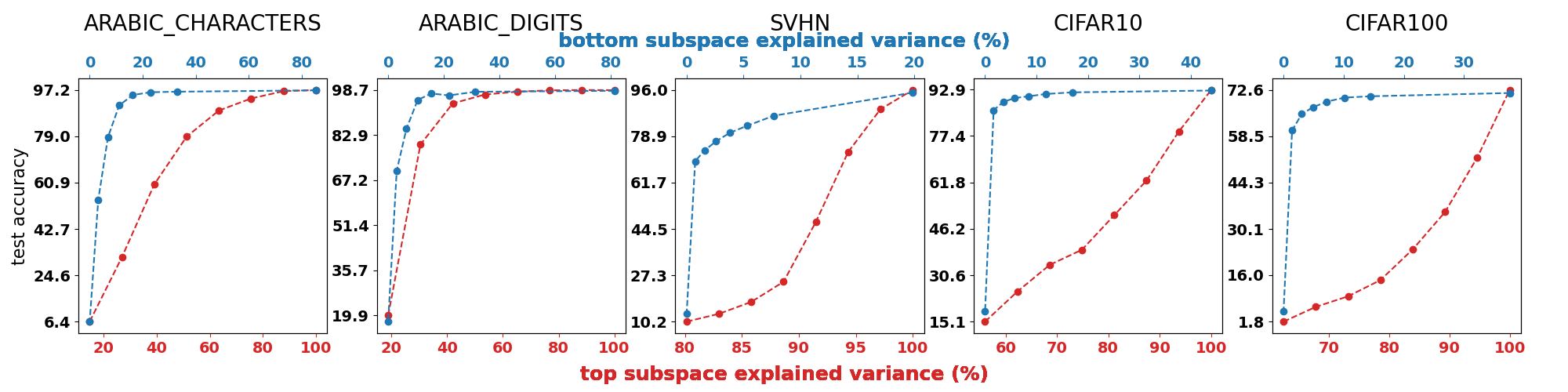}\\
    \caption{We depict the classification accuracy of a ResNet9 DNN when trained and tested on images that have been projected onto the top ({\bf red}) and bottom ({\bf blue}) subspace as ordered per the eigenvalues of the data covariance matrix, without data-augmentation ({\bf top}) and with data-augmentation ({\bf bottom}). We clearly observe that except for datasets without background and for which reconstruction and classification are better aligned (recall \cref{fig:alignment}), the final performance is greater when employing the subspace of the data that explains the least the pixel variation, i.e., the bottom subspace.}
    \label{fig:classification}
\end{figure*}

{\bf Linear regime results are informative.}~
We have focused on the linear encoder, decoder, and classification head of \cref{eq:bilinear}. Albeit insightful, one might wonder how much of those insights transfer to the more realistic setting of employing nonlinear mappings. We note that it remains common to keep the classification head linear therefore leading to the following generalization of \cref{eq:bilinear} as
\begin{align}
    \mathcal{L}(\mW\hspace{-0.04cm},\theta,\gamma) \hspace{-0.04cm}= &\| \mW^\top\hspace{-0.08cm} f_{\theta}(\mX)\hspace{-0.04cm} - \hspace{-0.06cm}\mY\|_F^2\hspace{-0.04cm}+\hspace{-0.04cm} \lambda\|g_{\gamma}\hspace{-0.04cm}\left(f_{\theta}(\mX)\hspace{-0.02cm}\right) \hspace{-0.04cm}- \hspace{-0.06cm}\mX \|_F^2.\label{eq:nonlinear}
\end{align}
The encoder is now the nonlinear mapping $f_{\theta}:\mathbb{R}^{D}\mapsto \mathbb{R}^{K}$ and the decoder is the nonlinear mapping $g_{\gamma}:\mathbb{R}^{K} \mapsto \mathbb{R}^{D}$. We formalize below a result that will reinforce the legitimacy of our linear regime analysis (\cref{thm:linear_solution,thm:condition,thm:alignment}) by showing that it is (i) a correct model during the early phase of training, and even (ii) a correct model throughout training when the decoder being employed is under-parametrized.

\begin{theorem}
\label{thm:linear_decoder}
For any high-capacity encoder $f_{\theta}$, studying \cref{eq:bilinear} and \cref{eq:nonlinear} is equivalent at initialization for any decoder, and is always equivalent when the decoder is linear. (Proof in \cref{proof:linear_decoder}.)
\end{theorem}

Combining \cref{thm:linear_decoder,thm:linear_solution}, we obtain that even with DNs, during the early stages of learning, the encoder-decoder mapping focuses on the principal subspace of the data, i.e., the space that explains most of the reconstruction error in the linear regime. As our study strongly hinges on that claim, we propose to empirically validate it in the following \cref{sec:misfit}.

\vspace{-0.2cm}
\subsection{Reconstruction and Perception Features Live In Different Subspaces of the Data}
\label{sec:misfit}
\vspace{-0.2cm}

We characterized in the previous \cref{sec:linear} how classification and reconstruction tasks fail to align when it comes to learning common features. In particular, \cref{sec:linear,fig:alignment} validated how training focuses first on the top subspace of the data. We now reinforce our claim by showing that supervised tasks can not be solved when restricting the images on the subspace that is learned first by reconstruction.

{\bf Perception can not be solved from the principal subspace of the data.}~We first propose a controlled experiment where we artificially remove some of the original data subspace. In particular, we consider two settings. First, we gradually remove the subspace associated to the top eigenvectors of the data covariance matrix effectively removing what is most useful for reconstruction but also what we claim to be least useful for perception. Second, we gradually remove the subspace associated to the bottom eigenvectors (the one least useful for reconstruction but that we claim to be most useful for perception). This procedure is applied to the entire dataset (train and test images) before any DN training occurs. Hence the DN is only presented with the filtered images. We report the top-1 accuracy over numerous datasets in \cref{fig:teaser} (top) and in \cref{fig:classification}. We obtain a few key observations. First, \;{\em for any \% of filtering, keeping the bottom subspace of the data produces higher test accuracy that when keeping the top subspace}. That is, the subspace that is most useful for reconstruction (top) is least useful for perception. Second, {\em the accuracy gap is impacted by the presence of background, finer-grained classes, and higher resolution images.} This further validates our theoretical observations from \cref{fig:alignment} and the result from \cref{thm:linear_decoder}. We will now focus on validating the second part of our claim that the subspace used for perception (bottom) is learned last, and slowly.

{\bf Useful features for perception are learned last.}~The above results demonstrate that the top subspace of the data--explaining most of the pixel variance--is not aligned with the perception tasks. Yet, perfect reconstruction implies capturing both the top and bottom subspace. Albeit correct, we demonstrate in \cref{fig:teaser} (bottom) and in \cref{fig:extra_focus} that the rate at which the bottom subspace is learned is exponentially slower than the rate at which the top subspace is learned. This empirically validates \cref{thm:linear_decoder}. For the reader familiar with optimization \cite{benzi2002preconditioning}, or power iteration methods, this observation is akin to how many procedures converge at a rate which is a function of the eigengap \cite{booth2006power,xu2018accelerated}, i.e., the difference between $\lambda_i$ and $\lambda_{i+1}$, where $\lambda$ are the sorted eigenvalues. Because natural images have an exponential decay of their eigenvalues \cite{van1996modelling,ruderman1997origins}, the rate at which the top subspace is approximated is exponentially faster than the bottom one, therefore making the learning of useful features for perception occur only late during training.
This finding also resonates with previous studies on the spectral bias of DNs in classification and generative settings \cite{chakrabarty2019spectral,rahaman2019spectral,schwarz2021frequency}.


Combining the observations from this section supports {\bf R1} and {\bf R2} from \cref{sec:intro}. It remains to study {\bf R3} which states that since features for perception lie within a negligible subspace (as measured by the reconstruction loss), one can find two separate models that equally solve the reconstruction task (same train and test loss values) but provide drastically different perception task performances.



\begin{figure*}[t!]
\begin{minipage}{0.24\linewidth}
    \includegraphics[width=\linewidth]{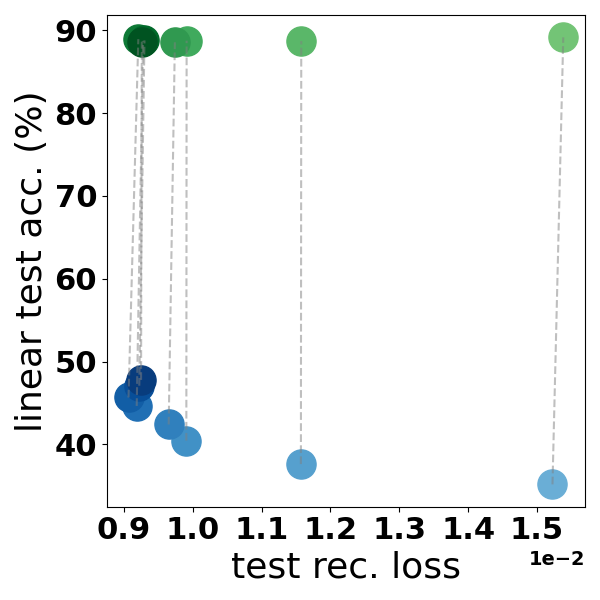}
\end{minipage}
\begin{minipage}{0.24\linewidth}
    \includegraphics[width=\linewidth]{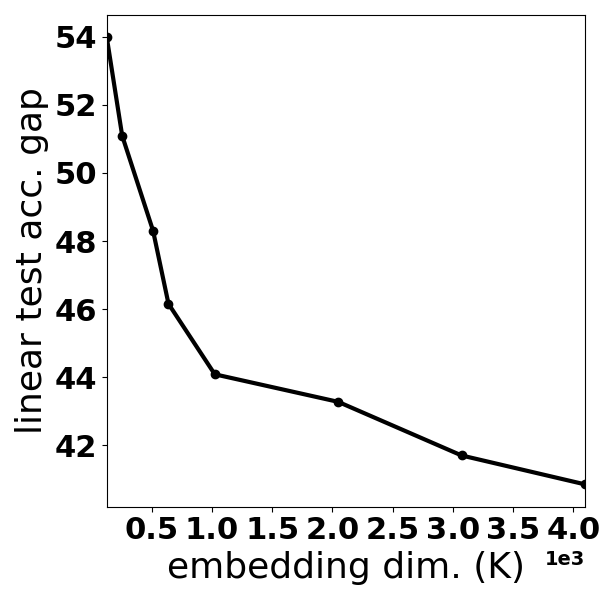}
\end{minipage}
\begin{minipage}{0.24\linewidth}
    \includegraphics[width=\linewidth]{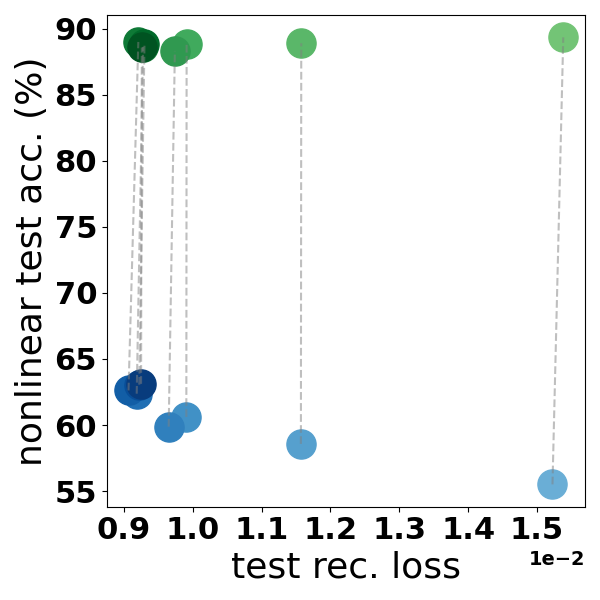}
\end{minipage}
\begin{minipage}{0.24\linewidth}
    \includegraphics[width=\linewidth]{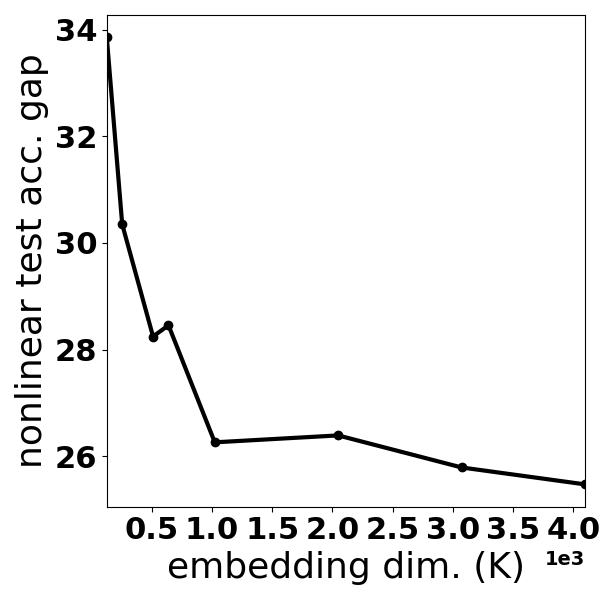}
\end{minipage}
\vspace{-0.2cm}
\caption{Depiction of multiple resnet34 autoencoders with varying embedding dimensions ({\bf light to dark}) some trained only to reconstruct the input samples with data-augmentations ({\bf blue}) and others with an additional supervised loss signal (as per \cref{eq:nonlinear}) ({\bf green}). We report the test set accuracy and the relative difference ({\bf y-axis}) for each of the ``paired'' models, i.e., the ones with every training setting identical except for the use of the supervised signal, as a function of the train and test rec loss. We clearly observe that for any embedding dimension and reconstruction loss, one can find two set of parameters with drastically different ability to solve perception tasks. Reconstructed samples and training curves are provided in \cref{fig:compare}.}
\label{fig:compare2}
\end{figure*}

\vspace{-0.2cm}
\section{Learning By Reconstruction Needs Guidance}
\label{sec:guidance}
\vspace{-0.2cm}

We now turn to {\bf R3} stating that features for perceptions lie within a space with negligible impact on the reconstruction error--therefore motivating the need to add additional guidance to the training, e.g., through denoising tasks. To do so, we will show that it is possible to obtain two DNs with same reconstruction error but one with perception capabilities far greater than the other (\cref{sec:guidance_1}). Lastly we will prove that some guidance can be provided to the learned representation to reduce that gap and focus towards more useful features through careful design of the denoising task (\cref{sec:DAE}).

\vspace{-0.2cm}
\subsection{Learning By Reconstruction Can Produce Optimal Representations}
\label{sec:guidance_1}
\vspace{-0.2cm}

\begin{figure*}[t!]
\fboxsep=1mm
\fboxrule=0mm
\fcolorbox{blue}{blue}{\includegraphics[width=\linewidth]{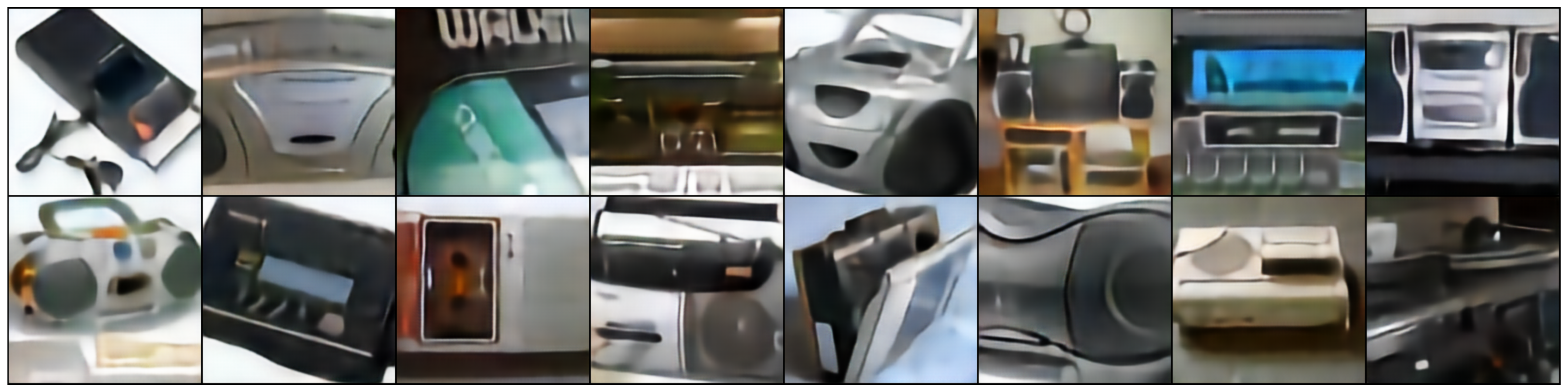}}\\
\fcolorbox{orange}{orange}{\includegraphics[width=\linewidth]{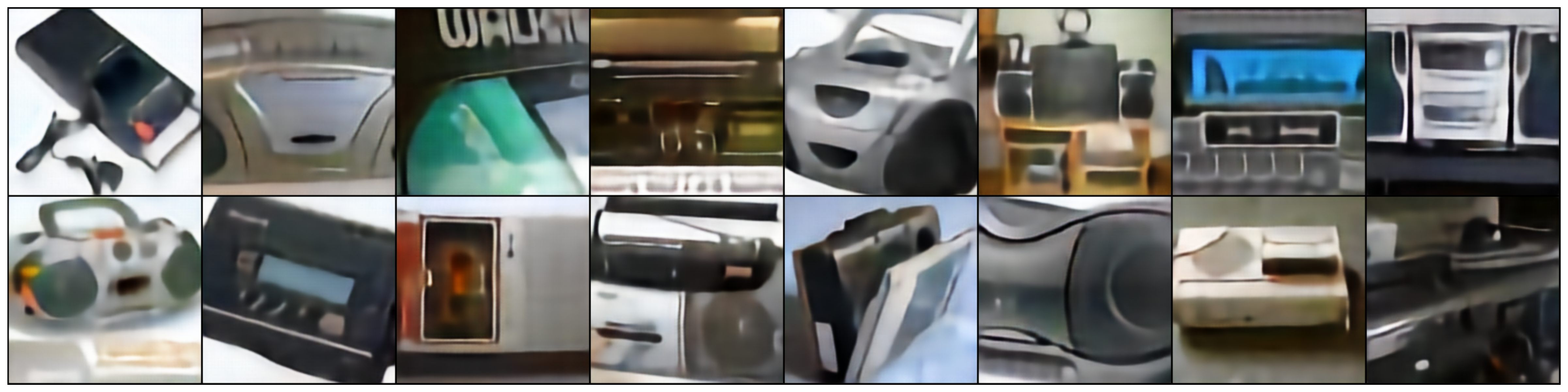}}\\
\includegraphics[width=\linewidth]{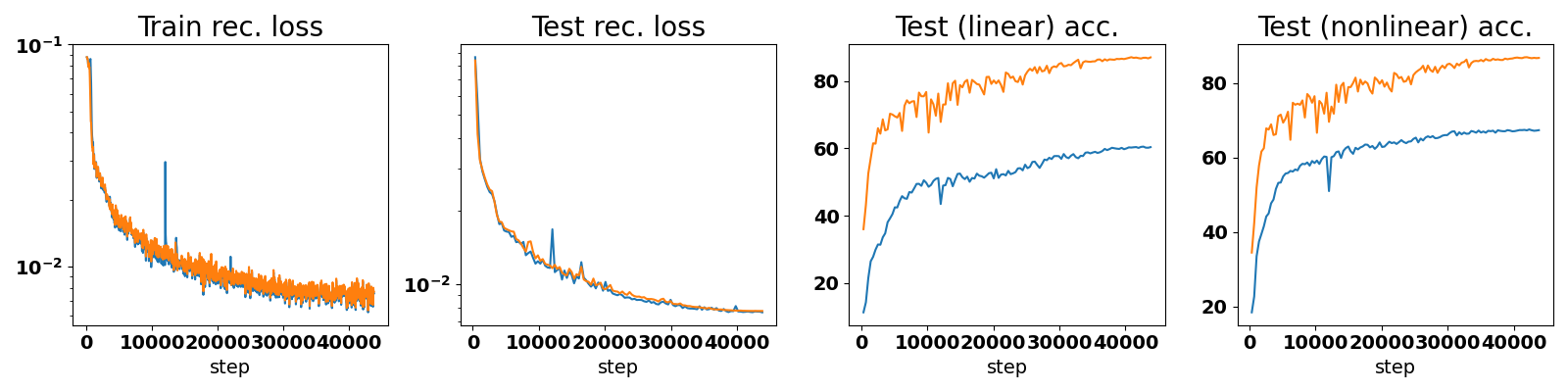}\\
    \vspace{-0.35cm}
\caption{\small Depiction of two resnet34 autoencoders trained on Imagenette (Imagenet-10) images, one (\color{orange}orange\color{black}) with an additional training signal that favors latent representations suited for classification, and the other (\color{blue}blue\color{black}) that is only the reconstruction loss. As per {\bf R1} and {\bf R2} the latter naturally focuses on suboptimal features as showcases in the test accuracy, both when using a linear or a nonlinear probe. Crucially, the autoencoder with the additional signal produces representations with much greater discriminative power in both the linear and nonlinear setting. Yet, and despite popular belief, doing so has no impact on the reconstruction losses on the train or test set, and thus no impact on the quality of the reconstruction presented at the {\bf top}. Therefore validating {\bf R3}.}
\label{fig:compare}
\end{figure*}

One interesting benefit emerging from the observations made for {\bf R1} and {\bf R2} is that guiding a DN to focus on the subspace containing informative features for perception has minimal impact on the reconstruction loss--as they focus on different subspaces. Therefore, we now propose a simple experiment to demonstrate the above argument. We take a resnet34 autoencoder train it with the usual reconstruction loss (MSE) on Imagenette. This gives us a model that (as per R1 and R2) fails to properly focus on discriminative feature. To obtain the second DN with improved classification performance, we simply add a classification head on top of the embedding of the encoder. That is, the same embedding that is fed to the decoder for reconstruction, is also fed to the linear classifier with supervised training loss (recall \cref{eq:nonlinear}). We obtain the key insight of {\bf R3} which is that one can produce two DNs with same training loss (reconstruction) and validation loss (reconstruction), but with significantly different classification performance, as reported in \cref{fig:compare2,fig:compare}.

To further understand that observation, we can recall the results from \cref{thm:linear_solution}. We demonstrated that the encoder ($\mV$) having $K$ dimensions at its disposal, is optimal when selecting the top singular vectors of the data matrix $\mX$. If $K$ is large enough that it encompasses both the top subspace of the data (which is learned first and has greatest impact on the reconstruction loss) and the bottom subspace of the data (which is useful for perception as per \cref{fig:teaser,fig:classification}), then both objective can coexist (recall \cref{thm:condition}) as long as enough capacity is given to the encoder. Therefore, we obtain the following key insight. {\em Whenever the capacity of the autoencoder is large, the encoder embedding can (and will at the end of training) include features useful for perception all while being able to reconstruct its inputs}. Again, for this to happen one requires the capacity of the encoder to grow with the image resolution (as more and more dimensions will be taken up by the top subspace), and with the complexity of the image background (again taking more dimensions in the top subspace) (recall \cref{sec:misfit}).

The above observation demonstrates that learning to reconstruct needs an additional training signal to focus towards discriminative features. As we will prove below, learning by denoising offers such a solution.

\vspace{-0.2cm}
\subsection{Provable Benefits of Learning by Denoising}
\label{sec:DAE}
\vspace{-0.1cm}

\begin{figure*}[t!]
    \centering
    \begin{minipage}{0.195\linewidth}
    \centering
    \includegraphics[width=\linewidth]{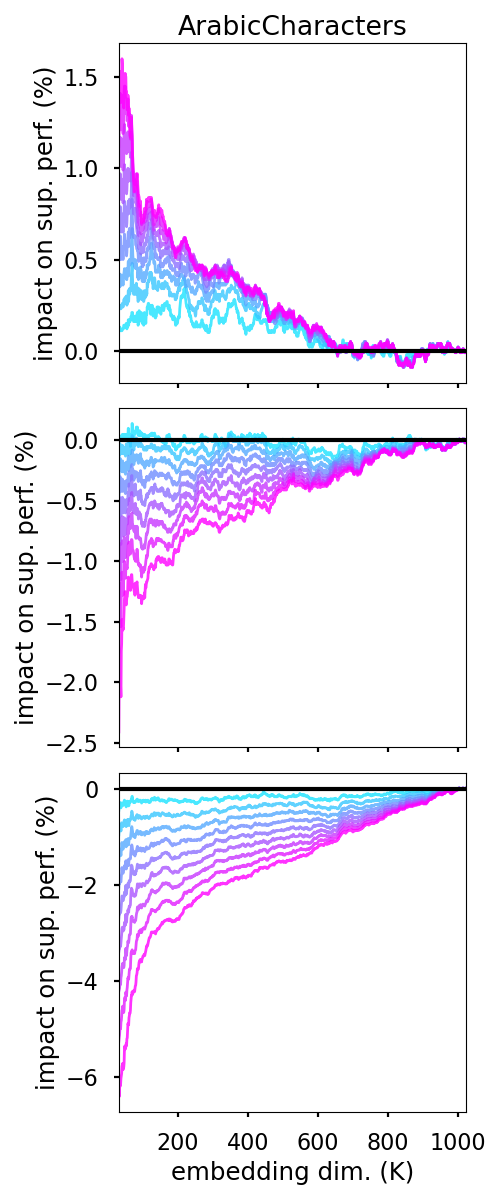}
    \end{minipage}
    \begin{minipage}{0.195\linewidth}
    \centering
    \includegraphics[width=\linewidth]{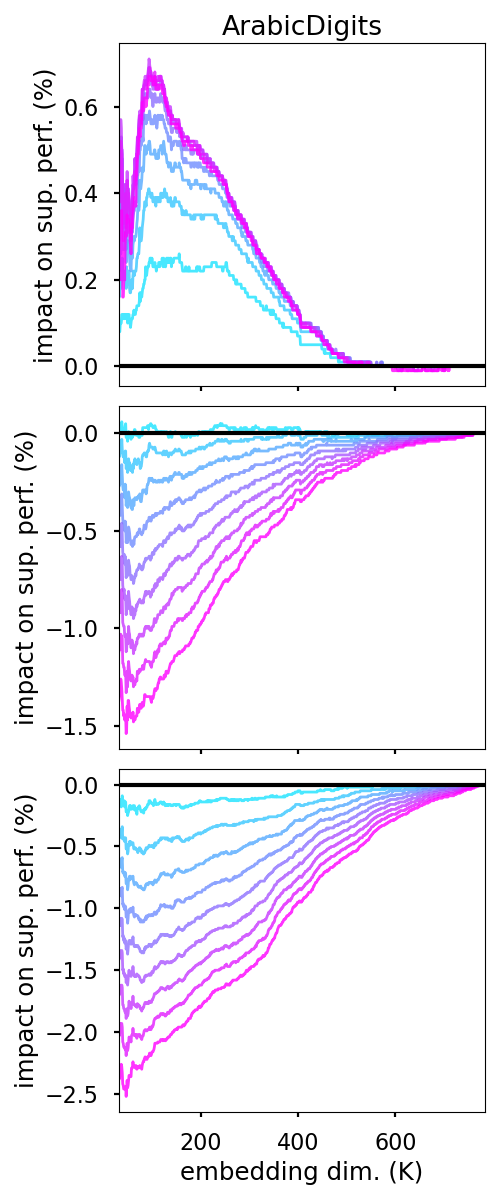}
    \end{minipage}
    \begin{minipage}{0.195\linewidth}
    \centering
    \includegraphics[width=\linewidth]{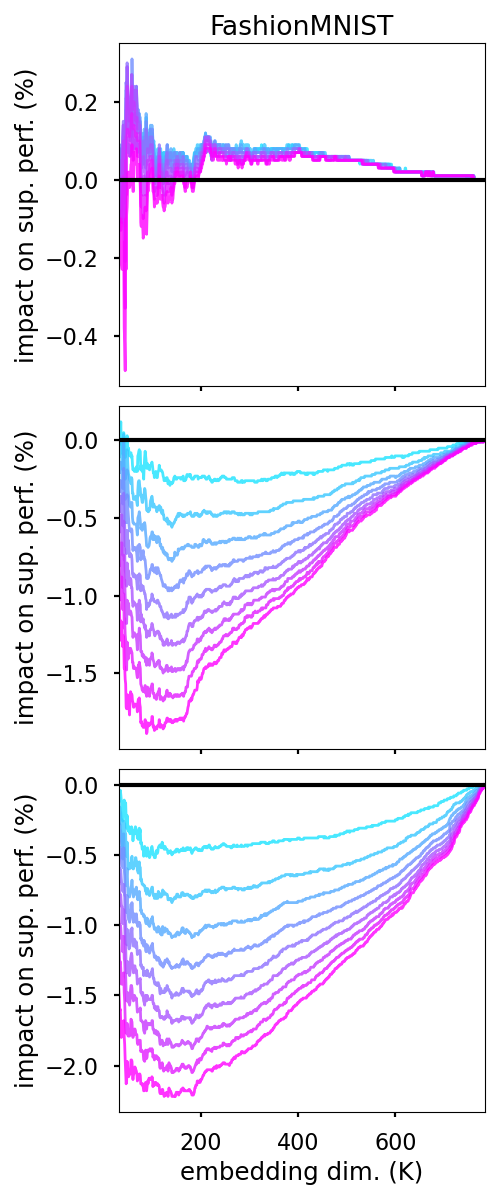}
    \end{minipage}
    \begin{minipage}{0.195\linewidth}
    \centering
    \includegraphics[width=\linewidth]{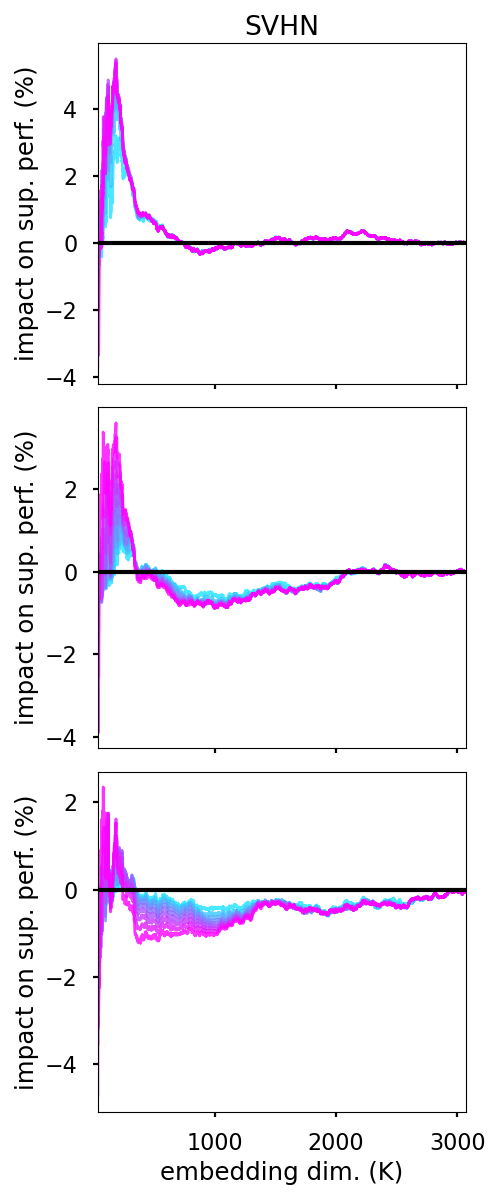}
    \end{minipage}
    \begin{minipage}{0.195\linewidth}
    \centering
    \includegraphics[width=\linewidth]{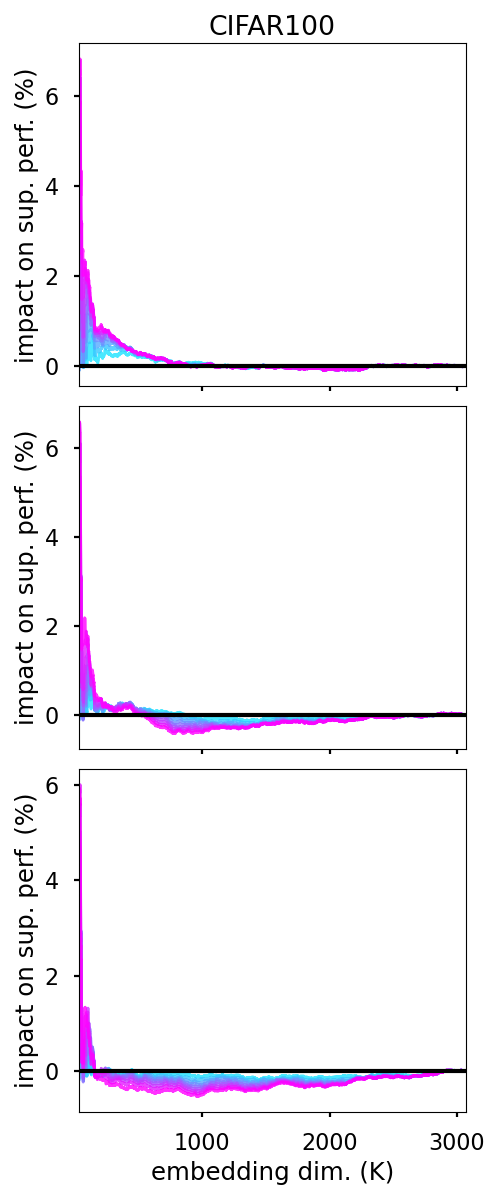}
    \end{minipage}
    \vspace{-0.2cm}
    \caption{Depiction of the relative alignment difference when employing denoising tasks (recall \cref{eq:optimal_V}) with masking noise, with probability of dropping ranging from $0\%$ to $99\%$ ({\bf cyan} to {\bf pink}) for patch size of $(1,1)$ recovering multiplicative dropout ({\bf top}), $(2,2)$ ({\bf middle}), and $(4,4)$ ({\bf bottom}) on various datasets. A positive number indicates a beneficial impact of using the denoising loss on the supervised performance of the learned representation. We observe that for datasets such as ArabicDigits that already have a strong alignment between the two tasks (recall \cref{fig:alignment}), the use of any form of masking is detrimental except with shape $(1,1)$. However for datasets such as CIFAR100 ({\bf right column}) with originally poor alignment, masking is beneficial and increases the alignment between the two tasks. As the original alignment increases with $K$, as the benefit of masking reduces.}
    \label{fig:alignment_expectation}
\end{figure*}

Recalling the Denoising Autoencoder setting from \cref{eq:DAE}, we aim to obtain a closed form solution of the linear loss \cref{eq:bilinear} in order to find some hints as to why masking and additive Gaussian noise produce representations of different quality for perception.

Our goal is therefore to study the misalignment metric (\cref{eq:alignment}) under the denoising setting which, as per \cref{thm:alignment}, is given by
\begin{align}
    \text{alignment}(k) \triangleq \min_{\mW}\| \mW^\top{\mV^*}^\top\mX - \mY\|_F^2,\nonumber\\
    {\mV^*} = \argmin_{\mV}\min_{\mZ} \mathbb{E}_{\mX'|\mX}\left[ \| \mZ^\top\mV^\top\mX'- \mX\|_F^2\right],\label{eq:optimal_V}
\end{align}
which is the minimum supervised loss that can be attained using the representation from $\mV^*$ that minimizes the denoising loss. We ought to highlight that we can obtain a closed form solution under the expectation over the noise distribution ($\mX'|\mX$) as formalized below, where we denote $G\triangleq \mathbb{E}_{\mX'|\mX}\left[\mX'{\mX'}^\top\right]$ and $\mS\triangleq \mathbb{E}_{\mX'|\mX}\left[\mX'\right]$.

\begin{theorem}
\label{thm:DAE}
The closed form solution for $\mV^*$ from \cref{eq:optimal_V} is given by $\mV^*  \text{ spans } \mP_{\mG}\mD^{-\frac{1}{2}}_{\mG}(\mP_{\mH})_{.,1:K}$,
where $\mH \triangleq  \mD^{-\frac{1}{2}}_{\mG}\mP^\top_{\mG}\mS\mX^\top\mX\mS^\top\mP_{\mG} \mD^{-\frac{1}{2}}_{\mG}$.
    (Proof in \cref{proof:DAE}.)
\end{theorem}

The above result demonstrates that even when employing denoising autoencoders with additive Gaussian noise or masking, we can obtain a closed form solution for $\mV$, and from that obtain all the alignment metrics studied so far. In particular, \cref{proof:DAE} also provides the closed form solutions for $\mG$ and $\mS$. We illustrate the alignment between reconstruction and perception tasks in the denoising autoencoder setting (\cref{eq:optimal_V}) in \cref{fig:alignment_expectation} for the case of random masking, as per the MAE setting. We clearly observe that the denoising task has the ability to increase the alignment between the two tasks--especially for small embedding dimension ($K$). We however observe that the size of the masked patches that provides the best gains vary with the dataset, hinging at another challenge of denoising autoencoders: the cross-validation of the denoising task. Another formal result we propose below will reinforce that point.

Denoting by $\mV^*(\sigma)$ the optimal denoising autoencoder parameters when employing additive isotropic Gaussian noise with standard deviation $\sigma$, we obtain the following statement showcasing that this type of denoising task does not help supervised tasks.

\begin{corollary}
\label{thm:gaussian_noise}
    Under the settings of \cref{thm:DAE}, additive Gaussian noise has no impact in the supervised task performance as ${\mW^*}^\top{\mV^*(\sigma)}^\top={\mW^*}^\top{\mV^*(0)}^\top,\forall \sigma \geq 0$, regardless of the supervised task. (Proof in \cref{proof:gaussian_noise}.)
\end{corollary}

We therefore obtain the following insights. {\em Denoising tasks offer a powerful guidance to skew learned representations to better align with perception tasks (\cref{fig:alignment_expectation}) but some noise distributions such as additive Gaussian noise are provably unable to help}. A challenge that naturally emerges is in selecting the adequate denoising tasks, e.g., to avoid \cref{thm:gaussian_noise} in a setting where labels are not available and the supervised tasks to be tackled may not be known a priori. An interesting portal that we obtained (\cref{thm:gaussian_noise}) in our study is that it is possible to assess if a denoising task has any impact on the perception task without yet knowing what is that supervised task. That alone could help in at least focusing on denoising tasks that do have an impact, albeit it will remain unknown if that impact will be beneficial or not.

\vspace{-0.2cm}
\section{Conclusion}
\vspace{-0.1cm}

We proposed to study the transferability of representations learned by reconstruction towards perception tasks. In particular, we obtain that the two objectives are fundamentally misaligned, with a degree of misalignment that grows with the presence of complicated background, with greater number of classes for the perception task, and with higher image resolutions. While our study focused on bringing those limitations forward from a theoretical and empirical angle, we also opened new avenues to reduce those limitations in the future. For example, we obtained a closed form solution to measure the impact of noise distributions to better align the learned representation to the downstream perception task. This novel methodology opens the door to a priori selecting noise distribution candidates. Even when the downstream task is unknown, we found that some noise distributions, such as additive Gaussian noise, are effectively unable to provide any benefit for better aligning reconstruction and perception tasks. On the opposite, we validated that masking is a valid strategy, albeit requiring some per-dataset tuning. That finding is in line with MAE's performances going from about 50\% to 74\% on Imagenet top-1 accuracy when masking is employed. We hope that our study will also open new avenues to study reconstruction methods for other modalities such as time-series and NLP.

\vspace{-0.2cm}
\section{Impact Statements}
\vspace{-0.2cm}
This paper presents work whose goal is to advance the field of Machine Learning. There are many potential societal consequences of our work, none which we feel must be specifically highlighted here.

{
    \small
    \bibliographystyle{icml2024}
    \bibliography{iclr2024_conference}
}

\onecolumn

\subsection{Proof of \cref{thm:linear_solution}}
\label{proof:linear_solution}

\begin{proof}
The first part of the proof finds the optimum $\mW^*$ and $\mZ^*$ as a function of $\mV$ which is direct since we are in a least-square style setting for each of them. The second part will consist in showing that the optimal $\mV$ can be found as the solution of a generalized eigenvalue problem. The third and final step will be to express the solution for $\mV$ in close-form that is also friendly for computations.

{\bf Step 1.}~Recall that our loss function is given by 
\begin{align*}
    \mathcal{L} = \| \mW^\top\mV^\top\mX-\mY\|_F^2 + \lambda \| \mZ^\top\mV^\top\mX-\mX\|_F^2,
\end{align*}
recalling that $\|\mM\|_F^2=\Tr(\mM^\top\mM)$, the above simplifies to
\begin{align*}
    \mathcal{L} =& \|\mY\|_F^2- 2 \Trp{\mX^\top\mV\mW\mY}+\Trp{\mX^\top\mV\mW\mW^\top\mV^\top\mX}\\
    &+ \lambda \|\mX\|_F^2 -2\lambda\Trp{\mX^\top\mV\mZ\mX} + \lambda\Trp{\mX^\top\mV\mZ\mZ^\top\mV^\top\mX},
\end{align*}
we are now going to find the optimal $\mW$ and $\mZ$ which are unique by convexity of the loss and of their domain. Recall that we assume $\mY$ and $\mX$ to be full-rank (therefore also making $\mV$ full-rank). Recalling the derive of traces, we obtain
\begin{align*}
    \nabla_{\mW}\mathcal{L} &= -2\mV^\top\mX\mY^\top + 2 \mV^\top\mX\mX^\top\mV\mW,\\
    \nabla_{\mZ}\mathcal{L} &= -2\lambda\mV^\top\mX\mX^\top + 2\lambda \mV^\top\mX\mX^\top\mV\mZ,
\end{align*}
setting it to zero (we assume here $\lambda>0$ otherwise we can not solve for $\mZ$ since its value does not impact the loss) and solving leads
\begin{align*}
    \mW^* =& (\mV^\top\mX\mX^\top\mV)^{-1}\mV^\top\mX\mY^\top,\\
    \mZ^* =& (\mV^\top\mX\mX^\top\mV)^{-1}\mV^\top\mX\mX^\top.
\end{align*}
We now have solved for $\mW,\mZ$ as a function of $\mV$, i.e., the loss is now only a function of $\mV$, which we are going to solve for now.

{\bf Step 2.}~We will first proceed by plugging the values for $\mW^*,\mZ^*$ back into the loss, which will now be only a function of $\mV$. Let's first simplify our derivations by noticing that
\begin{align*}
    \Trp{\mX^\top\mV\mW^*{\mW^*}^\top\mV^\top\mX}&=\Trp{\mX^\top\mV(\mV^\top\mX\mX^\top\mV)^{-1}\mV^\top\mX\mY^\top\mY\mX^\top\mV(\mV^\top\mX\mX^\top\mV)^{-1}\mV^\top\mX}\\
    &=\Trp{\mX^\top\mV(\mV^\top\mX\mX^\top\mV)^{-1}\mV^\top\mX\mY^\top\mY}\\
    &=\Trp{\mX^\top\mV\mW^*\mY},
\end{align*}
and similarly
\begin{align*}
    \Trp{\mX^\top\mV\mZ^*{\mZ^*}^\top\mV^\top\mX}=\Trp{\mX^\top\mV\mZ^*\mX},
\end{align*}
finally making the entire loss simplify as follows
\begin{align*}
    \mathcal{L} &= \|\mY\|_F^2- 2 \Trp{\mX^\top\mV\mW^*\mY}+\Trp{\mX^\top\mV\mW^*{\mW^*}^\top\mV^\top\mX}\\
    &\hspace{3cm}+ \lambda \|\mX\|_F^2 -2\lambda\Trp{\mX^\top\mV\mZ^*\mX} + \lambda\Trp{\mX^\top\mV\mZ^*{\mZ^*}^\top\mV^\top\mX}\\
    &=\|\mY\|_F^2- \Trp{\mX^\top\mV\mW^*\mY}+\lambda \|\mX\|_F^2 -\lambda\Trp{\mX^\top\mV\mZ^*\mX}\\
    &=\|\mY\|_F^2- \Trp{\mX^\top\mV    (\mV^\top\mX\mX^\top\mV)^{-1}\mV^\top\mX\mY^\top    \mY}\\
    &\hspace{2cm}+\lambda \|\mX\|_F^2 -\lambda\Trp{\mX^\top\mV (\mV^\top\mX\mX^\top\mV)^{-1}\mV^\top\mX\mX^\top  \mX}\\
    &=\|\mY\|_F^2+\lambda \|\mX\|_F^2 - \Trp{\mX^\top\mV (\mV^\top\mX\mX^\top\mV)^{-1}\mV^\top\mX\left(\mY^\top\mY + \lambda \mX^\top\mX\right)}.
\end{align*}
First, notice that both $\mX\left(\mY^\top\mY + \lambda \mX^\top\mX\right)\mX^\top$ and $\mX\mX^\top$ are symmetric. Therefore, we can minimize the loss by solving the following generalized eigenvalue problem:
\begin{align*}
    \text{ find $\mV \in \rmat{D}{D}$ so that } \mX\left(\mY^\top\mY + \lambda \mX^\top\mX\right)\mX^\top\mV = \vv^\top \mX\mX^\top \mV \Lambda,
\end{align*}
where $\mV$ are the eigenvectors of the generalized eigenvalue problem, and $\Lambda$ the eigenvalues, then the solution to our problem we be any rotation of any $K$ eigenvectors, but the minimum will be achieve for the top-K ones.

{\bf Step 3.} We will first demonstrate the general solution for the generalized eigenvalue problem. Given that solution, it will be easy to take the top-$K$ eigenvectors that solve the considered problem. Denoting $\mA \triangleq \mX\left(\mY^\top\mY + \lambda \mX^\top\mX\right)\mX^\top$ and $\mB \triangleq \mX\mX^\top$, and $\mH\triangleq \mD^{-\frac{1}{2}}_{\mB}\mP^\top_{\mB}\mA\mP_{\mB} \mD^{-\frac{1}{2}}_{\mB}$, we have
\begin{align*}
    \mA\mP_{\mB}\mD^{-\frac{1}{2}}_{\mB}\mP_{\mH}&=\mB\mP_{\mB}\mD^{-\frac{1}{2}}_{\mB}\mP_{\mH} \Lambda\\
    \iff  \mP_{\mH}^\top\mD^{-\frac{1}{2}}_{\mB}\mP_{\mB}^\top\mA\mP_{\mB}\mD^{-\frac{1}{2}}_{\mB}\mP_{\mH}&=\mP_{\mH}^\top\mD^{-\frac{1}{2}}_{\mB}\mP_{\mB}^\top\mB\mP_{\mB}\mD^{-\frac{1}{2}}_{\mB}\mP_{\mH} \Lambda&&(\mP_{\mH}^\top\mD^{-\frac{1}{2}}_{\mB}\mP_{\mB}^\top \text{ bijective})\\
    \iff  \mP_{\mH}^\top\mD^{-\frac{1}{2}}_{\mB}\mP_{\mB}^\top\mA\mP_{\mB}\mD^{-\frac{1}{2}}_{\mB}\mP_{\mH}&=\Lambda\\
    \iff  \mD_{\mH}&=\Lambda,
\end{align*}
therefore the eigenvalues are given by $\mD_{\mH}$ and the eigenvectors are given by $\mP_{\mB}\mD^{-\frac{1}{2}}_{\mB}\mP_{\mH}$ or equivalently $(\mP_{\mX\mX^\top}\mD^{-\frac{1}{2}}_{\mX\mX^\top}\mP_{\mH})_{.,1:K}$. So the optimal $\mV$ is any rotation of the top-$K$ eigenvectors. The above is simple to use as-is whenever $N >D$, if not, then we can obtain a solution without having to compute any $D \times D$ matrix, thus making the process more efficient. To that end, we can obtain
\begin{align*}
    \mM\triangleq \mY^\top \mY + \lambda \mX^\top \mX\\
    \mS \triangleq\mD_{\mM}^{\frac{1}{2}}\mP^\top_{\mM}\mX^\top\\
    \mP_{\mA} = \mV_{\mS}, \mD_{\mA}=\mSigma^2_{\mS},
\end{align*}
that only involves $D \times \min(D,N)$ matrices instead of $D \times D$.

\end{proof}

\subsection{Proof of \cref{thm:PCA}}
\label{proof:PCA}

\begin{proof}
    We will start with the fully supervised (least-square) proof obtained when $\lambda =0$.
    Also notice that in any case, we have that 
    \begin{align*}
        \mX = \mP_{\mX\mX^\top} \mD_{\mX\mX^\top}^{\frac{1}{2}}\mV_{\mX}^\top.
    \end{align*}

    {\bf Ordinary Least Square recovery.}~since $\lambda=0$ we also have
    \begin{align*}
        \mA =\mX\left(\mY^\top\mY + \lambda \mX^\top\mX\right)\mX^\top=\mX\mY^\top\mY\mX^\top
    \end{align*}
    when then lead to
    \begin{align*}
        \mH =& \mD^{-\frac{1}{2}}_{\mX\mX^\top}\mP^\top_{\mX\mX^\top}\mA\mP_{\mX\mX^\top} \mD^{-\frac{1}{2}}_{\mX\mX^\top}\\
        =& \mD^{-\frac{1}{2}}_{\mX\mX^\top}\mP^\top_{\mX\mX^\top}\mX\mY^\top\mY\mX^\top\mP_{\mX\mX^\top} \mD^{-\frac{1}{2}}_{\mX\mX^\top}\\
        =& \mD^{-\frac{1}{2}}_{\mX\mX^\top}\mP^\top_{\mX\mX^\top}\mP_{\mX\mX^\top} \mD_{\mX\mX^\top}^{\frac{1}{2}}\mV_{\mX}^\top\mY^\top\mY\mV_{\mX}\mD_{\mX\mX^\top}^{\frac{1}{2}}\mP_{\mX\mX^\top} ^\top\mP_{\mX\mX^\top} \mD^{-\frac{1}{2}}_{\mX\mX^\top}\\
        =& \mV_{\mX}^\top\mY^\top\mY\mV_{\mX},
    \end{align*}
    from the above, we can simply plug those values in the analytical form for $\mV^*$ from \cref{eq:V} to obtain
    \begin{align*}
        \mV=\mP_{\mX\mX^\top}\mD_{\mX\mX^\top}^{-\frac{1}{2}}\mP_{\mH}=\mP_{\mX\mX^\top}\mD_{\mX\mX^\top}^{-\frac{1}{2}}\mV_{\mX}^\top\mV_{\mY}=\mX^{\dagger}(\mV_{\mY})_{.,1:K}
    \end{align*}
    since we easily see that $\mP_{\mH}=\mV_{\mX}^\top\mV_{\mY}$. We also have the optimum for $\mW$ from \cref{eq:W} to be 
    \begin{align*}
        \mW=&({\mV^*}^\top\mX\mX^\top\mV^*)^{-1}{\mV^*}^\top\mX\mY^\top\\
        =&(\mV_{\mY}^\top\mV_{\mY})^{-1}\mV_{\mY}^\top\mV_{\mX}\mD_{\mX\mX^\top}^{-\frac{1}{2}}\mP_{\mX\mX^\top}^\top\mX\mY^\top\\
        =&\mV_{\mY}^\top\mV_{\mX}\mV_{\mX}^{\top}\mY^\top\\
        =&\mSigma_{\mY}\mU_{\mY}^\top
    \end{align*}
    and finally the product of both matrices (which produce the supervised linear model) is obtained as 
    \begin{align*}
        \mW^\top\mV^\top = \mU_{\mY}\mSigma_{\mY}({\mV_{\mY}}_{.,1:K})^\top(\mX^{\dagger})^\top=\mY\mX^\top(\mX\mX^\top)^{-1}, K\geq C,
    \end{align*}
    therefore recovering the OLS optimal solution. Note that if $K<C$ then we have a bottleneck and we therefore obtain an interesting alternative solution that looks at the top subspace of $\mY$ (this is however never the case in OLS settings).

    {\bf Principal Component Analysis recovery.}We now consider the case where we only employ the unsupervised loss (akin to $\lambda \to \infty$). In this setting we get
    \begin{align*}
        \mA=\mX\mX^\top\mX\mX^\top,
    \end{align*}
    and we directly obtain 
    \begin{align*}
        \mH=&\mD^{-\frac{1}{2}}_{\mX\mX^\top}\mP^\top_{\mX\mX^\top}\mA\mP_{\mX\mX^\top} \mD^{-\frac{1}{2}}_{\mX\mX^\top}\\
        =&\mD^{-\frac{1}{2}}_{\mX\mX^\top}\mP^\top_{\mX\mX^\top}\mX\mX^\top\mX\mX^\top\mP_{\mX\mX^\top} \mD^{-\frac{1}{2}}_{\mX\mX^\top}\\
        =&\mD_{\mX\mX^\top},
    \end{align*}
    therefore the optimal form for $\mV$ will be 
    \begin{align*}
        \mV=\mP_{\mX\mX^\top}\mD_{\mX\mX^\top}^{-\frac{1}{2}}(\mP_{\mH})_{.,1:K}=\mP_{\mX\mX^\top}(\mD_{\mX\mX^\top}^{-\frac{1}{2}})_{.,1:K},
    \end{align*}
    which will select the top-$K$ subspace of $\mX$ (recall that the eigenvalues of $\mH$ are $\mD_{\mX\mX^\top}$ and therefore its top-$K$ eigenvectors are selected the top-$K$ dimension of the subspace. Then the solution for $\mZ$ from \cref{eq:Z} gives
    \begin{align*}
        \mZ =& ({\mV^*}^\top\mX\mX^\top\mV^*)^{-1}{\mV^*}^\top\mX\mX^\top\\
        =&{\mV^*}^\top\mX\mX^\top\\
        =&((\mD_{\mX\mX^\top}^{\frac{1}{2}})_{.,1:K})^\top\mP_{\mX\mX^\top}^\top,
    \end{align*}
    and lastly the product of $\mZ$ and $\mV$ (which produce the final linear transformation processing $\mX$ takes the form
    \begin{align*}
        \mZ^\top\mV^\top=\mP_{\mX\mX^\top}(\mD_{\mX\mX^\top}^{\frac{1}{2}})_{.,1:K})((\mD_{\mX\mX^\top}^{-\frac{1}{2}})_{.,1:K})^\top\mP_{\mX\mX^\top}^\top=(\mP_{\mX\mX^\top})_{.,1:K}((\mP_{\mX\mX^\top})_{.,1:K})^\top,
    \end{align*}
    which is the projection matrix onto the top-$K$ subspace of the data $\mX$ i.e. recovering the optimal solution of Principal Component Analysis.
\end{proof}

\subsection{Proof of \cref{thm:alignment}}
\label{proof:alignment}

\begin{proof}
Recall that in the $\lambda \to \infty$ regime, we have that $\mV^*=\mP_{\mX\mX^\top}(\mD_{\mX\mX^\top}^{-\frac{1}{2}})_{.,1:K}$ and $\mW^*=\mP^\top_{\mX\mX^\top}\mY^\top$. We thus develop
    \begin{align*}
        \| {\mW^*}^\top{\mV^*}^\top\mX-\mY\|_F^2=&\|\mY\mP_{\mX\mX^\top}((\mD_{\mX\mX^\top}^{-\frac{1}{2}})_{.,1:K})^{\top} \mP_{\mX\mX^\top}^\top \mX - \mY \|_F^2\\
        =&\|\mY\mP_{\mX\mX^\top}((\mD_{\mX\mX^\top}^{-\frac{1}{2}})_{.,1:K})^{\top} \mP_{\mX\mX^\top}^\top \mP_{\mX\mX^\top} \mD_{\mX\mX^\top}^{\frac{1}{2}}\mP_{\mX\mX^\top}^\top - \mY \|_F^2\\
        =&\| \mY (\mP_{\mX\mX^\top})_{.,1:K}((\mP_{\mX\mX^\top})_{.,1:K})^\top-\mY\|_F^2\\
        =&\|\mY\|_F^2 - 2 \Trp{\mY (\mP_{\mX\mX^\top})_{.,1:K}((\mP_{\mX\mX^\top})_{.,1:K})^\top\mY^\top} + \| \mY (\mP_{\mX\mX^\top})_{.,1:K}((\mP_{\mX\mX^\top})_{.,1:K})^\top\|_F^2\\
        =&\|\mY\|_F^2 -  \Trp{\mY (\mP_{\mX\mX^\top})_{.,1:K}((\mP_{\mX\mX^\top})_{.,1:K})^\top\mY^\top}\\
        =&\|\mY\|_F^2 -  \|\mY (\mP_{\mX\mX^\top})_{.,1:K}\|_F^2,
    \end{align*}
as $\|\mY\|_F^2$ is a constant with respect to the parameters, we consider $\|\mY (\mP_{\mX\mX^\top})_{.,1:K}\|_F^2$ as our alignment measure (the greater, the better the supervised loss can be minimized from the parameters). Since this quantity lives in the range $[0,\|\mY \mP_{\mX\mX^\top}\|_F^2]$, we see that by using the reparametrization from \cref{eq:alignment} we obtain the proposed measure of alignment rescaled to $[0,1]$.
\end{proof}

\subsection{Proof of \cref{thm:DAE}}
\label{proof:DAE}

We need to find the optimal reconstruction solution $(\mV^*,\mZ^*)$ first (corresponding to the case of $\lambda \to \infty$, and then plug it into the supervised loss with optimal $\mW$.
\begin{align*}
    \mathbb{E}_{\mX'_n \sim p_{\mX'|\mX}}\| \mW^\top\mV^\top\mX' - \mX\|_2^2,
\end{align*}
from which we obtain
\begin{align*}
    \min_{\mW,\mV}\sum_{n=1}^{N}\mathbb{E}_{\vx'_n \sim p_{\vx'_n|\vx_n}}\Tr\left(\mW^\top\mV^\top\mX'\mX'^\top\mV\mW\right)-2\Tr\left(\mW^\top\mV^\top\mX'\mX^\top\right)+\cst,
\end{align*}
leading to $\mW^*=(\mV^{\top}\mathbb{E}[\mX'\mX'^\top]\mV)^{-1}\mV^\top\mathbb{E}[\mX']\mX^\top$ which we can plug back into the loss to obtain
\begin{align*}
    &\Tr\left(\mX\mathbb{E}[\mX']^\top\mV(\mV^{\top}\mathbb{E}[\mX'\mX'^\top]\mV)^{-1}\mV^\top\mathbb{E}[\mX'\mX'^\top]\mV(\mV^{\top}\mathbb{E}[\mX'\mX'^\top]\mV)^{-1}\mV^\top\mathbb{E}[\mX']\mX^\top\right)\\
    &-2\Tr\left(\mX\mathbb{E}[\mX']^\top\mV(\mV^{\top}\mathbb{E}[\mX'\mX'^\top]\mV)^{-1}\mV^\top\mathbb{E}[\mX']\mX^\top\right)+\cst,\\
    =&-\Tr\left(\mX\mathbb{E}[\mX']^\top\mV(\mV^{\top}\mathbb{E}[\mX'\mX'^\top]\mV)^{-1}\mV^\top\mathbb{E}[\mX']\mX^\top\right)+\cst,\\
    =&-\Tr\left(\mV^\top\mathbb{E}[\mX']\mX^\top\mX\mathbb{E}[\mX']^\top\mV(\mV^{\top}\mathbb{E}[\mX'\mX'^\top]\mV)^{-1}\right)+\cst,\\
\end{align*}
whose solution is given (assuming $\mathbb{E}[\mX'\mX'^\top]$ is full-rank) by the solution of the generalized eigenvalue problem
\begin{align*}
    \argmax_{\vv} \frac{\vv^\top \mathbb{E}[\mX']\mX^\top\mX\mathbb{E}[\mX']^\top \vv}{\vv^\top \mathbb{E}[\mX'\mX'^\top] \vv}.
\end{align*}

Given those optimum we can thus obtain the alignment measure same as before as the sueprvised loss obtain from $V^*$ from the unsupervised loss and $\mZ^*$ from the supervised one:
\begin{align*}
   \mZ^*=\argmin_{\mZ} \mathbb{E}_{\mX'_n \sim p_{\mX'|\mX}}\| \mZ^\top{\mV^*}^\top\mX' - \mY\|_2^2,
\end{align*}
whose optimum is therefore given by $\mZ^*=({\mV^*}^{\top}\mathbb{E}[\mX'\mX'^\top]{\mV^*})^{-1}{\mV^*}^\top\mathbb{E}[\mX']\mY^\top$ which can then be plugged back to produce the (unnormalized) measure. The analytical form of can be obtained (following the derivations of \cite{balestriero2022data}) as follows for example for the case of additive Gaussian noise which is trivial and commonly done before e.g. showing the link between ridge regression and additive dropout $\mathbb{E}[\mX']=\mX$ and $\mathbb{E}[\mX'\mX'^\top]=\mX'\mX'^\top + \sigma \mI$. The perhaps more interesting derivations concern the masking as employed by MAE.

\subsection{Proof of \cref{thm:gaussian_noise}}
\label{proof:gaussian_noise}

In the case of additive, centered Gaussian noise, we have $\mathbb{E}[\mX']=\mX$ and $\mathbb{E}[\mX'\mX'^\top]=\mX'\mX'^\top + \sigma \mI$. Therefore the optimal value for $\mV$, is given by solving the generalized eigenvalue problem $(\mX\mX^\top\mX\mX^\top,\mX\mX^\top + \sigma \mI)$.
Recalling the derivations of the optimal solution for such problem from \cref{proof:linear_solution}, we have that 
\begin{align*}
    \mV = \mP_{\mX\mX^\top + \sigma \mI}\mD^{-\frac{1}{2}}_{\mX\mX^\top + \sigma \mI}(\mP_{\mH})_{.,1:K}=\mP_{\mX\mX^\top}(\mD_{\mX\mX^\top} + \sigma \mI)^{-\frac{1}{2}}(\mP_{\mH})_{.,1:K},
\end{align*}
with 
\begin{align*}
    \mH =& \mD^{-\frac{1}{2}}_{\mX\mX^\top + \sigma \mI}\mP^\top_{\mX\mX^\top + \sigma \mI}(\mX\mX^\top\mX\mX^\top)\mP_{\mX\mX^\top + \sigma \mI}\mD_{\mX\mX^\top + \sigma \mI}^{-\frac{1}{2}}\\
    =& \mD^{-\frac{1}{2}}_{\mX\mX^\top + \sigma \mI}\mP^\top_{\mX\mX^\top}(\mX\mX^\top\mX\mX^\top)\mP_{\mX\mX^\top}\mD_{\mX\mX^\top + \sigma \mI}^{-\frac{1}{2}}\\
    =& (\mD_{\mX\mX^\top})^2(\mD_{\mX\mX^\top} + \sigma \mI)^{-1},
\end{align*}
and the important property to notice is that the ordering of the eigenvalues of $\mH$ which are given by $(\mD^\top_{\mX\mX^\top})^2(\mD_{\mX\mX^\top} + \sigma \mI)^{-1}$ are the same as the ordering of $\mX\mX^\top$ which are given by $\mD^\top_{\mX\mX^\top}$. That is, the top-$K$ subspace that will be picked up by $\mP_{\mH}$ are the same for any noise standard deviation $\sigma$. Now given the close form for $\mV$ we can obtain the close form of the classifier weights $\mW$ from \cref{eq:W} to be 
    \begin{align*}
        \mW=&({\mV^*}^\top\mX\mX^\top\mV^*)^{-1}{\mV^*}^\top\mX\mY^\top\\
        \mW=&\left(((\mP_{\mH})_{.,1:K})^\top (\mD_{\mX\mX^\top} + \sigma \mI)^{-\frac{1}{2}}\mP_{\mX\mX^\top}^\top\mX\mX^\top\mP_{\mX\mX^\top}(\mD_{\mX\mX^\top} + \sigma \mI)^{-\frac{1}{2}}(\mP_{\mH})_{.,1:K}\right)^{-1}\\
        &\times ((\mP_{\mH})_{.,1:K})^\top (\mD_{\mX\mX^\top} + \sigma \mI)^{-\frac{1}{2}}\mP_{\mX\mX^\top}^\top\mX\mY^\top\\
        \mW=&\left(((\mP_{\mH})_{.,1:K})^\top (\mD_{\mX\mX^\top} + \sigma \mI)^{-1}\mD_{\mX\mX^\top}(\mP_{\mH})_{.,1:K}\right)^{-1}((\mP_{\mH})_{.,1:K})^\top (\mD_{\mX\mX^\top} + \sigma \mI)^{-\frac{1}{2}}\mD_{\mX\mX^\top}^{\frac{1}{2}}\mV_{\mX}^\top\mY^\top\\
        \mW=&((\mP_{\mH})_{.,1:K})^\top (\mD_{\mX\mX^\top} + \sigma \mI)\mD_{\mX\mX^\top}^{-1}(\mP_{\mH})_{.,1:K}((\mP_{\mH})_{.,1:K})^\top (\mD_{\mX\mX^\top} + \sigma \mI)^{-\frac{1}{2}}\mD_{\mX\mX^\top}^{\frac{1}{2}}\mV_{\mX}^\top\mY^\top\\
        \mW=&((\mP_{\mH})_{.,1:K})^\top \left((\mD_{\mX\mX^\top} + \sigma \mI)^{\frac{1}{2}}\mD_{\mX\mX^\top}^{-\frac{1}{2}}\right)_{s}\mV_{\mX}^\top\mY^\top,
    \end{align*}
    where the $s$ subscript indicates that only the top $K \times K$ part of the diagonal matrix is nonzero.
    Lastly, the product of both matrices (producing the supervised linear model) is obtained as 
    \begin{align*}
        \mW^\top\mV^\top =& \mY \mV_{\mX} \left((\mD_{\mX\mX^\top} + \sigma \mI)^{\frac{1}{2}}\mD_{\mX\mX^\top}^{-\frac{1}{2}}\right)_{s}(\mP_{\mH})_{.,1:K}((\mP_{\mH})_{.,1:K})^\top (\mD_{\mX\mX^\top} + \sigma \mI)^{-\frac{1}{2}}\mP_{\mX\mX^\top}^\top\\
        =& \mY \mV_{\mX} (\mD_{\mX\mX^\top}^{-\frac{1}{2}})_{s}\mP_{\mX\mX^\top}^\top
    \end{align*}
    therefore recovering the OLS optimal solution $\mY\mX^\top(\mX\mX^\top)^{-1}$ whenever $K\geq D$, and otherwise recovers the projection onto the top subspace of $\mX$--in any case the final parameters are invariant to the choice of the standard deviation of the additive Gaussian noise ($\sigma$) during the denoising autoencoder pre-training phase.

\subsection{Proof of \cref{thm:linear_decoder}}
\label{proof:linear_decoder}

\begin{proof}
The first of the proof is to rewrite the joint objective classification and reconstruction objective with an arbitrary encoder network $f_{\theta}$
\begin{align*}
    \min_{\theta \in \mathbb{R}^{P},\mW \in \rmat{C}{K},\mV \in \rmat{D}{K}}\| \mW f_{\theta}(\mX)-\mY \|_F^2 + \lambda \| \mV f_{\theta}(\mX)-\mX\|_F^2,
\end{align*}
as the nonparametric version
\begin{align*}
    \min_{\mZ \in \rmat{K}{N},\mW \in \rmat{C}{K},\mV \in \rmat{D}{K}} \| \mW\mZ-\mY \|_F^2 + \lambda \| \mV\mZ-\mX\|_F^2,
\end{align*}
both being identical if we assume that the encoder is powerful enough to reach any representation, which is a realistic assumption given current architectures. Given that nonparametric objective, we can now solve for both the optimal decoder weight $\mV$ and the optimal representation $\mZ$ as follows
\begin{align*}
&\min_{\mZ \in \rmat{K}{N},\mW \in \rmat{C}{K},\mV \in \rmat{D}{K}} \| \mW\mZ-\mY \|_F^2 + \lambda \| \mV\mZ-\mX\|_F^2\\
    =&\min_{\mZ \in \rmat{K}{N}} \| \mY\mZ^\top(\mZ\mZ^\top)^{-1}\mZ-\mY \|_F^2 + \lambda \| \mX\mZ^\top(\mZ\mZ^\top)^{-1}\mZ-\mX\|_F^2\\
    =&\min_{\mZ \in \rmat{K}{N}} \Trp{\mZ^\top(\mZ\mZ^\top)^{-1}\mZ\mY^\top\mY\mZ^\top(\mZ\mZ^\top)^{-1}\mZ}-2\Trp{\mY^\top\mY\mZ^\top(\mZ\mZ^\top)^{-1}\mZ} + \|\mY\|_F^2\\
    &+\lambda\Trp{\mZ^\top(\mZ\mZ^\top)^{-1}\mZ\mX^\top\mX\mZ^\top(\mZ\mZ^\top)^{-1}\mZ}
    -2\lambda\Trp{\mX^\top\mX\mZ^\top(\mZ\mZ^\top)^{-1}\mZ} + \lambda\|\mX\|_F^2\\
     =&\min_{\mZ \in \rmat{K}{N}} \Trp{(\mZ\mZ^\top)^{-1}\mZ\mY^\top\mY\mZ^\top}-2\Trp{\mY^\top\mY\mZ^\top(\mZ\mZ^\top)^{-1}\mZ} + \|\mY\|_F^2\\
    &+\Trp{(\mZ\mZ^\top)^{-1}\mZ\mX^\top\mX\mZ^\top}
    -2\lambda\Trp{\mX^\top\mX\mZ^\top(\mZ\mZ^\top)^{-1}\mZ} + \lambda\|\mX\|_F^2\\
     =&\min_{\mZ \in \rmat{K}{N}} -\Trp{\mY^\top\mY\mZ^\top(\mZ\mZ^\top)^{-1}\mZ} + \|\mY\|_F^2
    -\lambda\Trp{\mX^\top\mX\mZ^\top(\mZ\mZ^\top)^{-1}\mZ} + \lambda\|\mX\|_F^2\\
      =&\min_{\mZ \in \rmat{K}{N}} -\Trp{\left(\mY^\top\mY+\lambda\mX^\top\mX\right)\mZ^\top(\mZ\mZ^\top)^{-1}\mZ} + \|\mY\|_F^2 + \lambda\|\mX\|_F^2\\
      =&\min_{\mZ \in \rmat{K}{N}: \mZ^\top\mZ = \mI} -\Trp{\left(\mY^\top\mY+\lambda\mX^\top\mX\right)\mZ^\top\mZ} + \|\mY\|_F^2 + \lambda\|\mX\|_F^2
\end{align*}
which is solved by $\mZ$ being any orthogonal matrix in the subspace of the top-K eigenvectors of $\left(\mY^\top\mY+\lambda\mX^\top\mX\right)$. Now as $\lambda \to \infty$ as the encoder will become more and more linear, ultimately converging to $f_{\theta}(\vx) = \mU\vx$ with $\mU\in \spn \{\eigvec(\mX^\top\mX)_1,\dots,\eigvec(\mX^\top\mX)_K\}$ 
\end{proof}
\subsection{Eigendecomposition}
\label{sec:fast_eigen}

Given a matrix $\mX \in \rmat{D}{N}$ with $D > N$, computing the eigendecomposition of $\mX \mX^\top$, a $D \times D$ matrix is $\mathcal{O}(D^3)$ which instead can be obtained in $\mathcal{O}(N^3 + DN^2)$ as
\begin{minipage}{\linewidth}
\begin{lstlisting}[language=Python,escapechar=\%]
def fast_gram_eigh(X, major="C", unit_test=False):
    """
    compute the eigendecomposition of the Gram matrix:
    - XX.T using column (C) major notation
    - X.T@X using row (R) major notation
    """
    if major == "C":
        X_view = X.T
    else:
        X_view = X

    if X_view.shape[1] < X_view.shape[0]:
        # this case is the usual formula
        U, S = np.linalg.eigh(X_view.T @ X_view)
    else:
        # in this case we work in the tranpose domain
        U, S = np.linalg.eigh(X_view @ X_view.T)
        S = X_view.T @ S
        S[U>0] /= np.sqrt(U[U>0])
        # ensuring that we have the correct values
        if unit_test:
            Uslow, Sslow = np.linalg.eigh(X_view.T @ X_vew)
            assert np.allclose(U, Uslow)
            assert np.allclose(S, Sslow)
    return U, S
\end{lstlisting}
\end{minipage}
since we have the relation
\begin{align*}
    \mX\mX^\top\vv=\lambda \vv \iff \mX^\top\mX(\mX^\top\vv) = \frac{\lambda}{\|\mX^\top\vv\|_2^2}(\mX^\top\vv),
\end{align*}
and thus we can simply compute the eigenvectors of the $K \times K$ matrix $\mX^\top\mX$ and get the eigenvectors of the $N\times N$ matrix $\mX \mX^\top$  by left-multiplying them by $\mX^\top$, and their corresponding eigenvalues are rescaled by $\frac{1}{\|\mX^\top\vv\|_2^2}$.

\subsection{Fast Implementation of Alignment Metric (\cref{eq:alignment})}
\label{sec:fast_alignment}

We want to sweep over the latent dimension $K$. As such, we can avoid recomputing the metric for each value and get them all at once as below. We again use the column-major notations as per \cref{sec:background}:
\begin{lstlisting}[language=Python,escapechar=\%]
def alignment_sweep(X, Y, major="C"):
    U, S, Vh = np.linalg.svd(X, full_matrices=False)
    if major == "C":
        denom = np.square(np.linalg.norm(Y @ Y.T))
        numer = np.linalg.multi_dot([Y.T, Y, Vh.T])
    else:
        denom = np.square(np.linalg.norm(Y.T @ Y))
        numer = np.linalg.multi_dot([Y,Y.T, U])
    numer = np.linalg.norm(numer, axis=0)**2
    return np.cumsum(numer) / denom
\end{lstlisting}

\subsection{Additional figures}

\begin{figure*}[h]
    \centering
    \begin{minipage}{0.03\linewidth}
    \rotatebox{90}{ bottom 25\% \hspace{5.3cm} top 75\%\hspace{5.3cm}original}
    \end{minipage}
    \begin{minipage}{0.8\linewidth}
    \includegraphics[width=\linewidth]{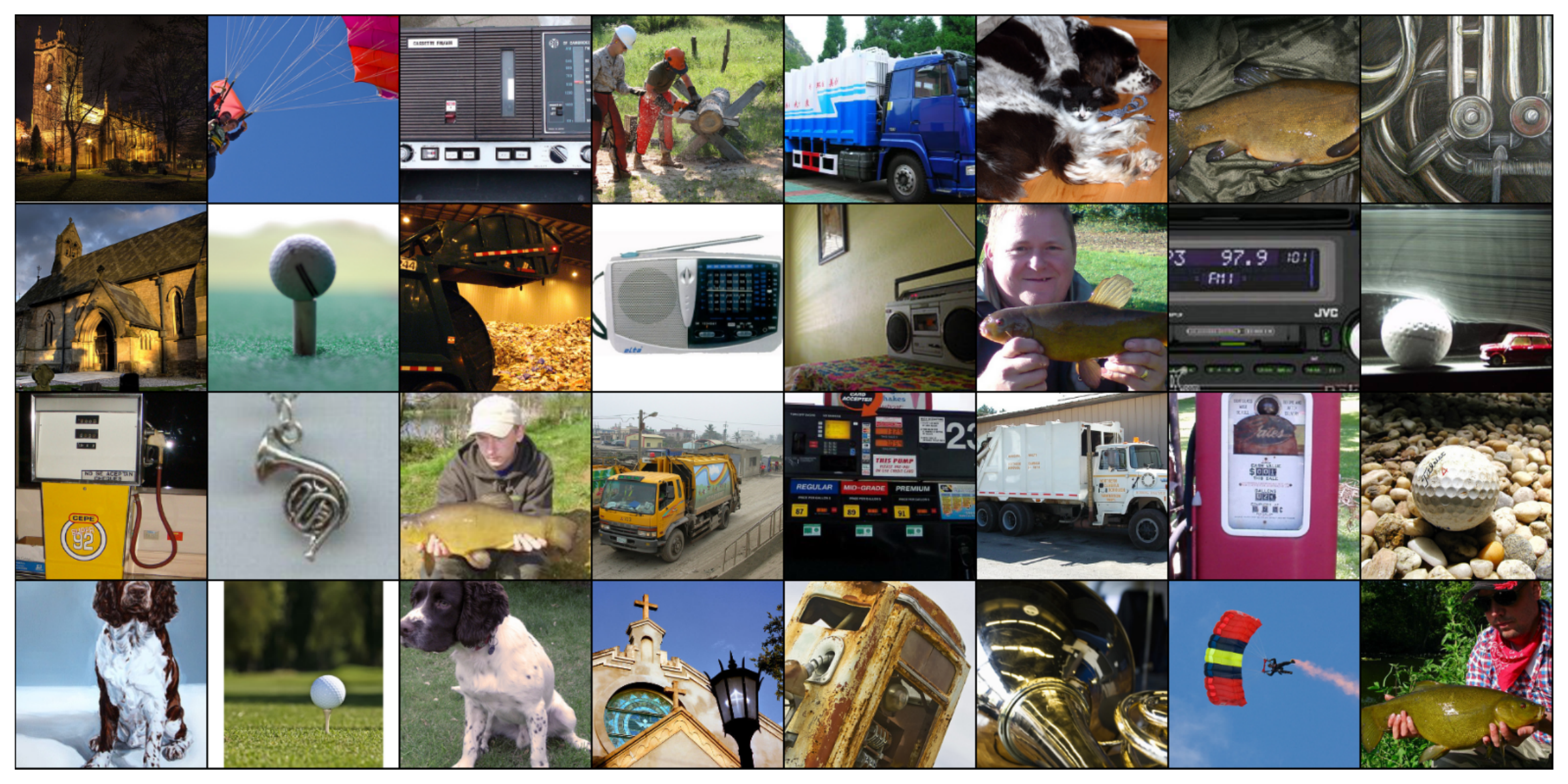}\\
    \includegraphics[width=\linewidth]{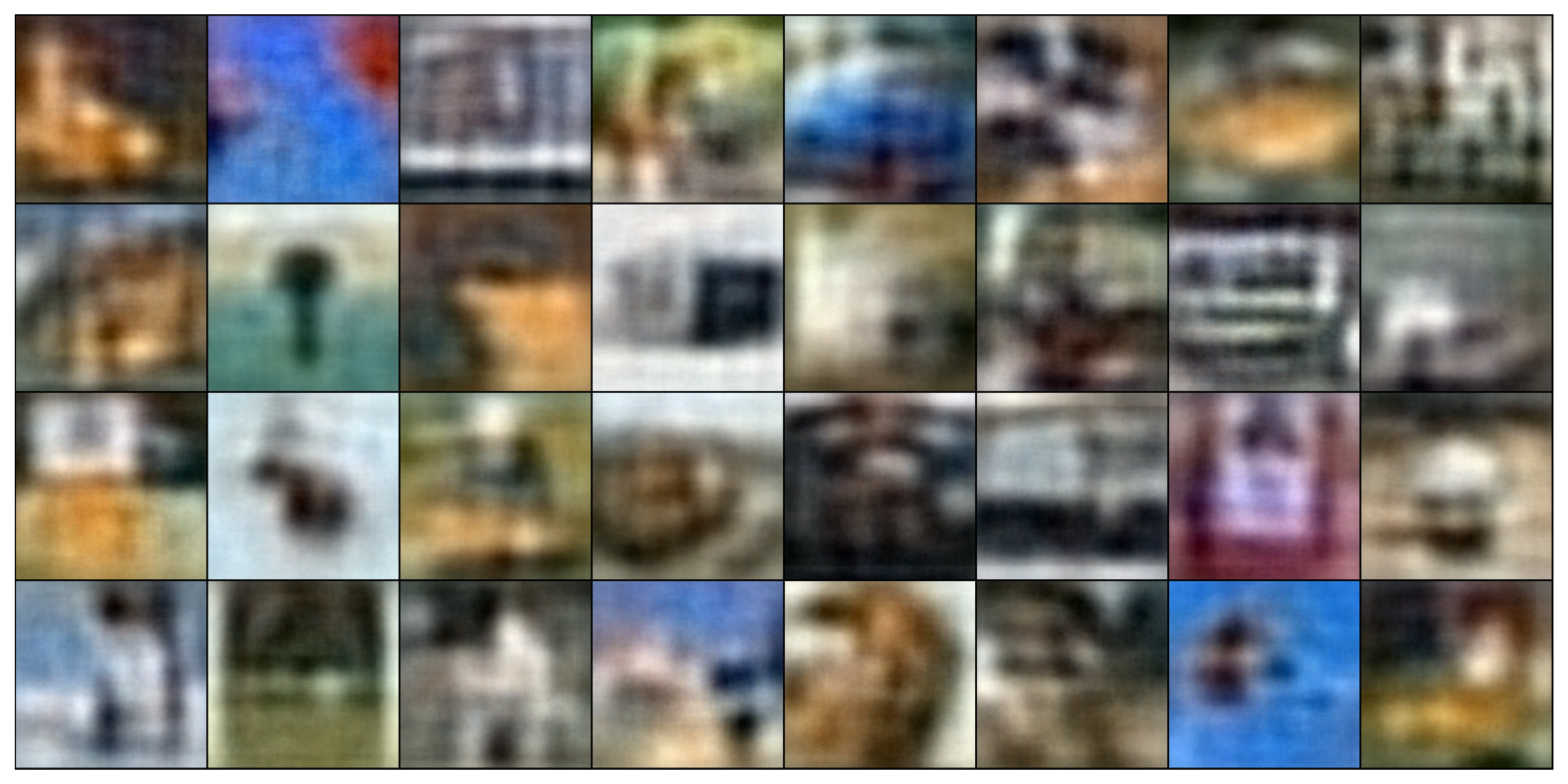}\\
    \includegraphics[width=\linewidth]{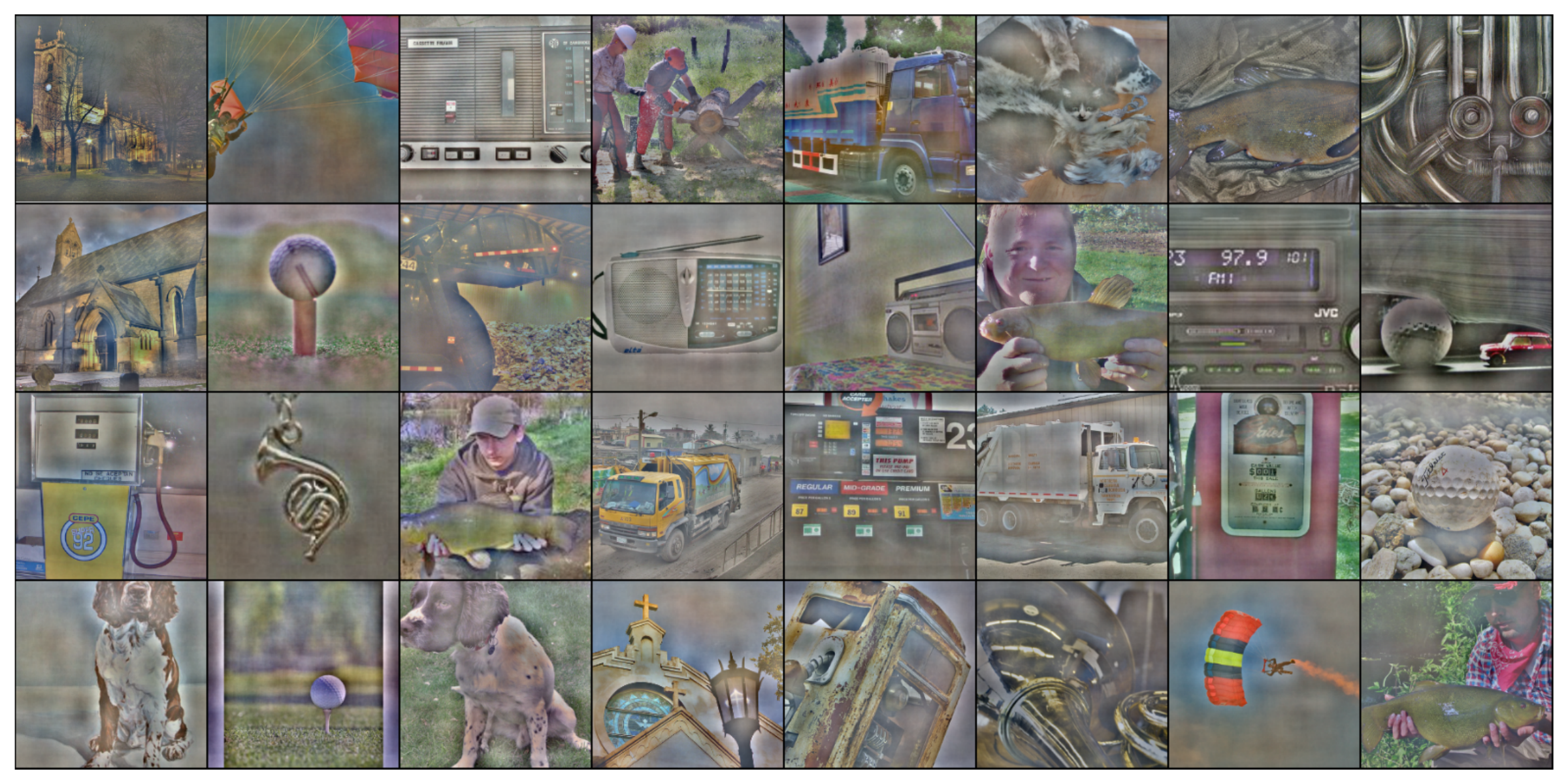}
    \end{minipage}
    \caption{Depiction of Imagenet images ({\bf top}) projected onto different subspaces obtained from Principal Component Analysis corresponding to the subspace explaining the top 75\% of pixel variance ({\bf middle}) and bottom 25\% of pixel variance ({\bf bottom}). We clearly observe that the image representation preserved after projection onto the bottom subspace makes the perception tasks (classification) easier to solve that if projected onto the top subspace, where the lower frequency information is insufficient to classify what is the object depicted (recall the classification performances of DNs applied onto those different projections from \cref{fig:teaser,fig:classification}).}
    \label{fig:images_pca}
\end{figure*}

\bibliography{iclr2024_conference}
\bibliographystyle{iclr2024_conference}

\appendix
\section{Appendix}
You may include other additional sections here.

\end{document}